%% file: main.tex
\pdfoutput=1

\documentclass[11pt]{article}

\usepackage{ACL2023}

\usepackage{times}
\usepackage{latexsym}

\usepackage[T1]{fontenc}

\usepackage[utf8]{inputenc}

 \usepackage{amsmath}

\usepackage{microtype}

\usepackage{inconsolata}

\usepackage{multirow}
\usepackage{hhline}
\usepackage{makecell}
\usepackage{hyperref}

\usepackage{arydshln}

\usepackage[]{graphicx}
\usepackage{subfigure}

\usepackage[utf8]{inputenc}
\usepackage[english]{babel}

\title{Age Recommendation from Texts and Sentences for Children}

\author{Rashedur Rahman \\ Univ Rennes, CNRS, IRISA \\  Lannion, France \\ \texttt{md-rashedur.rahman@irisa.fr}
        \And
        Gw\'enol\'e Lecorv\'e \\ Orange \\ Lannion, France \\ \texttt{gwenole.lecorve@orange.com}
        \AND
        Nicolas B\'echet \\ Univ. Bretagne Sud, CNRS, IRISA \\ Vannes, France \\ \texttt{nicolas.bechet@irisa.fr}}

\begin{document}
\maketitle
\vspace{7mm}
\begin{abstract}
Children have less text understanding capability than adults. Moreover, this capability differs among the children of different ages.
Hence, automatically predicting a recommended age based on texts or sentences would be a great benefit to propose adequate texts to children and to help authors writing in the most appropriate way.
This paper presents our recent advances on the age recommendation task. 
We consider age recommendation as a regression task, and discuss the need for appropriate evaluation metrics, study the use of state-of-the-art machine learning model, namely Transformers, and compare it to different models coming from the literature. Our results are also compared with recommendations made by experts.  
Further, this paper deals with preliminary explainability of the age prediction model by analyzing various linguistic features. We conduct the experiments on a dataset of $3,673$ French texts (132K sentences, 2.5M words). To recommend age at the text level and sentence level, our best models achieve MAE scores of $0.98$ and $1.83$ respectively on the test set. Also, compared to the recommendations made by experts, our sentence-level recommendation model gets a similar score to the experts, while the text-level recommendation model outperforms the experts by an MAE score of $1.48$.
\end{abstract}

\input{section_1_introduction}
\input{section_2_related_work}

\input{section_3_task}
\input{section_4_data}
\input{section_5_metrics}

\input{section_6_prediction_experiments_new}
\input{section_7_feature_analysis_EN_v3}
\input{section_8_conclusion}

\section*{Acknowledgement}

This work has been partially funded by the French National Research Agency (ANR) as part of the TextToKids project (ANR-19-CE38-0014).

\bibliography{main}
\bibliographystyle{acl_natbib}

\appendix
\section*{Appendix}
\input{appendix_metrics}

\input{appendix_expert_features}
\input{appendix_results}

\end{document}

%% file: section_1_introduction.tex
\section{Introduction}

Nowadays, children spend considerable time on the Internet. The primary concern is to ensure them to accessing adequate online content depending on their age. Therefore, in recent years, safe Internet for children has gained interest in many research domains~\citep{tomczyk2016children,byrne2017children,livingstone2019eu}.
Most studies focus on filtering abusive texts containing hate, violence, pornography, etc.~\citep{liu2015text,suvorov2013method}.
However, the adequacy also relates to the capacity of a child to understand the textual content (blogs, newspapers, etc.). Beyond the issue of safely browsing the web, this question of comprehensibility more widely concerns all types of documents that could be submitted or narrated to children, like books, schoolbook, etc. Thus, besides the impact on the user experience (children's and parent's ones), recommending and understanding the linguistic adequacy of a textual content for children of a given age is of primary interest to help children build up their knowledge.

Overall, text understanding by children is a well known topic in psycho-linguistics and cognitive sciences~\citep{Gathercole_1999,Tartas2010,mouw2019contributions}.
In particular, key findings have shown the impact of memory~\citep{Gathercole_1999}, temporality~\citep{Tartas2010,hickmann2012diversite}, and emotions~\citep{davidson2006role,mouw2019contributions}. On the contrary, this is still poorly studied in Natural Language Processing (NLP). Yet, the automatic recommendation of a target age for a given text could be benefit to various applications. For instance, this could help search engines to provide custom search settings by indexing a recommended age along with each document/web page. Computer-assisted writing tools could also be proposed to help authors or editors to propose suitable texts.

Toward this objective, this paper studies the task of automatically predicting a recommended age from textual productions. This task is addressed as a machine learning regression problem where multiple dimensions are defined and studied. The idea is to provide a solid and detailed frame for further progress in this area, both regarding methodological aspects and experimental results. This includes the following contributions:
\begin{enumerate}
\item The task of age recommendation from texts is defined and discussed to clarify difficulties or shortcomings possibly induced by some key choices, especially when defining the the fuzzy notion of "recommended" age. This mixes questions about the quality/reliability of the data annotations, the biases that specific text genres/styles can introduce or the evaluation of the results.
\item The paper proposed a dataset for textual age recommendation. This dataset is composed of French texts of various genres (novels, newspapers, and encyclopedias) for a wide range of ages. Moreover, the approach for the construction of this dataset can be easily transposed to other languages. 
\item  A particular attention is paid to the potentially relevant metrics for the task through the proposition of various metrics. Their respective behavior is studied through an artificial study case, and the best metrics are then used all along the paper.
\item With existing machine learning techniques, various recommendation models are proposed and compared. They cover various settings such that age can be recommended from texts or isolated sentences; using word/sentence embeddings, linguistic expert features or both; and with more or less complex models (transformers, feed-forward neural network or random forests). The results are compared against naive, histical and expert baselines. 
\item Since explainability is required for the perspective of human interactive applications, the paper proposes an opening where various linguistic features are studied to understand what makes a text easy or difficult to understand.
\end{enumerate}

In the remainder, Section~\ref{sec:related_work} discusses the related work, Section~\ref{sec:task} precisely defines the age recommendation task, and  Section~\ref{sec:data} details the dataset used for the experiments. An exhaustive study of the evaluation metrics is provided in Section~\ref{sec:metrics} which is followed by the recommendation models and experimental results in Section~\ref{sec:age_pred_experiments}. Finally, Section~\ref{sec:feature_analysis} investigates which linguistic features contribute most for age recommendation.

%% file: section_2_related_work.tex
\section{Related Work}
\label{sec:related_work}
Age recommendation from text is related to text understanding/readability analysis, and it has been studied in the fields of human and social sciences (Section~\ref{sec:human_sciences}), and NLP (Section~\ref{sec:nlp}).

\subsection{Human and Social Sciences}
\label{sec:human_sciences}
In the literature of psycho-linguistics and cognitive sciences, text comprehension by children has been studied well. Children have less text understanding capabilities than adults as their brain is still developing. The brain activity of a child peaks at 4~years. For language acquisition, the activity can be equivalent to 150\% of an adult~\citep{Gathercole_1999}. Short-term memory affects language understanding or accomplishment of complex tasks, and such memory is developed particularly between two to eight years~\citep{Gathercole_1999}. Acquisition of temporal notions is crucial for children to understand calendar-time and chronological orders~\citep{Tartas2010,hickmann2012diversite}. 
Emotions are also important factors to establishing and maintaining the coherency
of facts in a text~\citep{mouw2019contributions}, and the basic emotions (joy, anger, sadness, fear, etc.) are acquired around the age
of 10~\citep{davidson2006role}. When children start learning to read, \cite{frith1985beneath} argued that reading is acquired through three main stages: logographic, alphabetical, and orthographic stage. The logographic stage refers to the faculty of recognizing the drawing of a word rather than deciphering it, and is developed between the ages of 5 to 6 years. At later ages, the second stage (alphabetical), splitting a word into simpler graphical units (graphemes) and phonological units (phonemes), and the third stage (orthographic), linguistic ability to break down a word into meaningful units (morphemes) are developed.

However, one can relate age recommendation to the task of text readability, where various scores have historically been proposed to quantify the difficulty of a text, like
 the Flesch-Kincaid score~\citep{FleshKincaidGradeScore}, Dale-Chall formula~\citep{dale1948formula,chall1995readability} or the Gunning fog index~\citep{robert1968technique}. The Flesch-Kincaid score computes frequency ratios on syllables, while the Dale-Chall formula additionally counts difficult words. In contrast, the Gunning fog index considers the ratio of word to sentence counts, and the ratio of complex words to total words counts in a text. 
 More recently, \cite{todirascu2013coherence} studied the use of coherence and cohesion properties to assess the readability of texts, and they found some good correlations between such property and readability.
 \cite{islam2014readability_children} proposed to predict texts into four classes (very easy, easy, medium, and difficult) where the readability classes corresponded to some child age ranges. The approach inherits the readability index features along with the lexical features.
 In a very recent work, \cite{wilkens2022fabra} developed a readability assessment toolkit for french texts that aggregates various linguistic variables to train a classifier and a regression model. Also, this toolkit makes a ranking of the different features by measuring their correlations to the 9 readability levels.

\subsection{Natural Language Processing}
\label{sec:nlp}
In spite of these psycho-linguistic studies, the adequacy between ages and linguistic or cognitive skills has been less studied in the field of NLP. 
The first step comes from~\cite{ReadingLevelAssesment} where the authors explored how to automatically predict from which US school grade a child could read news articles. This problem was considered as a classification task among four classes. The model was a support vector machine with word n-gram, lexical and syntactic features. 

The traditional study of text readability~\citep{dell2011read,islam2012text} or age recommendation~\citep{ReadingLevelAssesment} mostly relies on the hand-crafted linguistic features which may become very difficult to extract for low-resource languages.
In recent years, using neural networks for text readability classification~\citep{mesgar2018neural,balyan2020applying,martinc2021supervised,feng2022cnn} and age prediction from text~\citep{bayot2017age,chen2019joint} has risen great interest in the field of computational linguistics. One major strength of the current approaches is to delegate the step of feature extraction to the neural network model based on the sole word embeddings of the input texts. Hence, such methods do not anymore necessarily require hand-crafted features for the training purpose.
For instance, \citet{blandin2020age} explored feed-forward neural networks to recommend the age corresponding to a text. 
The authors showed, among a large set of various linguistic features, word embedding features were the most contributory.
However, these models still suffer to achieve very good scores, and lack explainability of the predictions, but they show how difficult the task is. 

Apart from age recommendation or text readability, one popular related task is to characterize authors of textual contents. In \citet{nguyen2011author}, linear regression was used with gender and different textual features to predict the age of the authors of blogs, telephone conversations, and online forum posts. 
%
Similarly, \citet{chen2019joint} proposed LSTM-based regression and classification models to guess adult ages (from 14 to 34 years) of blog authors. The same task is performed by~\citet{bayot2017age} on tweets using a Convolutional Neural Network (CNN). Finally, an active learning based approach has also been explored in \citet{chen2016active} as a regression task for age prediction from social media texts by using textual and social features.
The task of author's age prediction from text is also a difficult one as the prediction performances are not so good yet. However, these studies show prospects of using regression models for age recommendation task.

Finally, in a very broad scope,  in the last years, transformers~\citep{wolf2019transformers} have become a very attractive neural architecture for many NLP tasks. Especially, encoder models like BERT~\citep{devlin2019bert} for many regression and classification tasks. Generally, such a model is pre-trained with a very large dataset and then it is used to perform on some downstream tasks by fine-tuning it with a smaller dataset. This architectural improvement motivates us to explore the advanced models for our age recommendation task. Nonetheless, it is interesting to highlight that these models are slower and more energy-consuming than former machine learning models. Hence, their benefit in terms of performance for the task should be balanced with the relative benefit compared to simpler approaches.

Yet, to our knowledge, no research studied any pre-trained BERT type model for age recommendation from text. In this paper, we propose to use CamemBERT~\citep{martin-etal-2020-camembert}, a pre-trained French language model built on the RoBERTa~\citep{DBLP:journals/corr/abs-1907-11692} architecture. We consider age recommendation as a regression task. To compare with our proposed CamemBERT approaches, this paper explores also Feed-forward and a Random Forest models. 
Moreover, we introduce two metrics (by tackling some special cases) for evaluating age recommendation, and provide a deeper analysis on these metrics. Finally, to explain the recommendations or predictions (as a preliminary expalinability), this paper presents different feature ranking approaches to show which linguistic features are most contributory for age recommendation.

%% file: section_3_task.tex
\section{Task Definition}
\label{sec:task}

We aim to assist children in their reading activities (online, leisure, or in schools) by suggesting them comprehensible texts according to their ages, and to facilitate the authors in writing the appropriate texts for the children. Therefore, the goal is to predict the target age from a given text.

\subsection{Challenges}
Text understanding capability differs to different children based on their ages, literacy, vocabulary etc. The children of same age may not have the same capacity of comprehending a piece of text. Some related studies worked on recommending texts to different school-grade levels~\cite{ReadingLevelAssesment}, and on text-complexity classification for different groups of school-grade levels~\cite{islam2014readability}. However, the grade levels corresponding to child-ages vary in different cultures and/or countries. When some grade levels or child-ages are merged into a group (or class), another question arises how large the interval of grade levels (or child-ages) is to be allowed. On the contrary, if one considers text recommendation for each individual child-age (or grade level), it lacks sufficient annotated data (to train a recommendation model) corresponding to the ages. 
Thus, from the computational linguistic point of view, defining the target age for automatically recommending age is very difficult. It requires to do some trade-offs among the different issues.

\subsection{Target Age}
In age recommendation task, the target age can be defined in several ways. 
First, one can think of the minimum age at which a text is comprehensible. However, in real world, suggesting a text to a child based on the minimum age may not be very practical as he/she cannot find it very interesting after certain age. For example, in general, a child of 12 years does not like the text (poem, fiction etc.) written for the kids of 4 years although he/she understands it very well. Such issues can be tackled by defining the target age as an interval (or range) of lower bound and upper bound, $[a, b]$ meaning that a text, $T$ is comprehensible by the children of ages between $a$ and $b$ years. This interval further can be deduced to a mean-age, $\mu=\frac{a + b}{2}$ with a considerable deviation. For example, if a text corresponds to the age-range of $[4-8]$, the target age is $6$ years with a deviation of $2$ years. In this paper, the target age is mainly defined as an age-range, but it studies the mean-age($\mu$) as well. 


\subsection{Recommendation Method}
Regarding the definition of the target age, we consider age recommendation as a regression task to predict the age as a numeric value from an input. The input of age recommendation can be either a full text or a sentence. 
The text level and sentence level recommendations allow us to deeper analyze different aspects and to serve different objectives of this task. 
In text level age recommendation method, the target age is predicted by analyzing the full text as a whole. However, for sentence level recommendation, each sentence is analyzed separately. Further, the sentence level predictions can be aggregated (e.g., mean aggregation) to make a text level global recommendation. Thus, the text level recommendation method can serve only the readers while the sentence level recommendation method can be useful for an author as well to incrementally verify (while writing a text) if the text is suitable for the readers of a certain age.

In age recommendation, an input text/sentence is represented with several features which are computed by various linguistic analysis. The feature-values are fed to a regression model in a supervised manner. Such a model is able to predict an age-range, an individual bound (min or max), and a mean age depending on the training setup.

%% file: section_4_data.tex
\section{Data}
\label{sec:data}

This section presents the corpus of $3,673$ French texts ($132$K sentences, $2.5$M words) that we have constituted to perform age recommendation. The main statistics of this corpus are given in Table~\ref{table:dataStat} and are described in the remainder. This corpus is intended to be publicly available\footnote{Due to the legal issue, now we are currently able to publicly release 25\% of this dataset. The complete dataset will be released very soon.}.

\newcommand{\range}[2]{[~#1~,~#2~]}

\begin{table*}[t!]
\centering
    \small
	\renewcommand{\arraystretch}{1.3}
	\begin{tabular}{|c|r|c|c|c|c|}
	 \hline
	  & \textbf{Genre}  & \textbf{\# texts}  & \textbf{\# sentences}  & \textbf{Age range (avg.)}  & \textbf{Mean age (avg.)}  \\
	  \hline
	  \hline
\multirow{4}{*}{Train} & Encyclopedia & 738 & 21,960 & \range{11.33}{15.71} & 13.52 \\
                         & Newspaper & 793 & 18,245 & \range{09.44}{13.85} & 11.64  \\
                         & Fiction & 966 & 47,277 & \range{09.06}{11.85} & 10.45 \\
\hhline{~-----}
                         & \textbf{Overall} & \textbf{2,509} & \textbf{87,882} & \textbf{\range{09.73}{13.26}} & \textbf{11.49} \\
\hline
\hline

\multirow{4}{*}{Validation} & Encyclopedia & 201 & 6,495 & \range{10.62}{15.13} & 12.87  \\
                         & Newspaper & 147 & 3,224 & \range{09.38}{13.78} & 11.58 \\
                         & Fiction & 255 & 12,471 & \range{07.86}{10.90} & 09.38 \\
\hhline{~-----}
                         & \textbf{Overall} & \textbf{605} & \textbf{22,261} & \textbf{\range{08.90}{12.57}} & \textbf{10.74} \\
\hline
\hline

\multirow{4}{*}{Test} & Encyclopedia & 135 & 4,325 & \range{11.52}{15.86} & 13.69 \\
                         & Newspaper & 189 & 4,118 & \range{09.49}{13.86} & 11.67 \\
                         & Fiction & 231 & 13,186 & \range{08.66}{11.83} & 10.24 \\
\hhline{~-----}
                         & \textbf{Overall} & \textbf{559} & \textbf{21,701} & \textbf{\range{09.41}{13.03}} & \textbf{11.22} \\
\hline
\hline

\multirow{4}{*}{Total} & Encyclopedia & 1074 & 32,780 & \range{11.21}{15.61} & 13.41 \\
                         & Newspaper & 1129 & 25,587 & \range{9.44}{13.84} & 11.64 \\
                         & Fiction & 1452 & 72,934 & \range{8.78}{11.68} & 10.23 \\
\hhline{~-----}
                         & \textbf{Overall} & \textbf{3,673} & \textbf{131,844} & \textbf{\range{9.53}{13.11}} & \textbf{11.32} \\
\hline
    \end{tabular}
\caption{Sizes and ages for the whole, training, validation and tests datasets. Statistics are given overall and for each text genre.}
\label{table:dataStat} 
\end{table*}

\subsection{Age Annotations}

All texts are associated with an age range. As previously discussed, this range is interpreted as the rough interval of ages from which the text is mostly adequate and can be fully understood. Age ranges come from indications given by the authors or editors in the case of children-dedicated texts, whereas they are arbitrary set to \range{14}{18} for adults.
It is important to highlight that these annotations are probably imperfect since editors and authors are not psycho-linguists\footnote{Experiments of Section~\ref{sec:experts} provide an idea of the differences between these reference age ranges and those recommended by experts.}. Indeed, age ranges are usually empirically defined, based on the feeling and experience of the authors and editors, as well as marketing considerations\footnote{For instance, age ranges for novels/stories are usually associated as a whole to collections, rather than individually to each book. Furthermore, some strategies may incite to lower the age range if this can help differenting a product from its concurent. Furthermore, age ranges tend to be large to attract more parents.}.

When working on isolated sentences later on in the paper, please note that all sentences will be associated with age range of the text from which they come.
For example, if a text annotated with an age range of $[a, b]$, all the sentences in the text are considered with the same age range. This is an inaccurate assumption since it is clear that all sentences of a difficult text may not be difficult when taken in isolation. Maybe only few of them are difficult or even the difficulty may come from the reasoning linking the sentences. However, this is the best approximation that one could reasonably expect as annotating thousands of sentences would be difficult and there would not be any guarantee that these annotations are consistent with the text level ones (again, the logics of the editors and authors is unknown).

\subsection{Balancing and partitioning}

To avoid linguistic biases due to specific genres, the corpus mixes encyclopedic, newspaper and fictional texts.
Encyclopedic texts comes from Vikidia\footnote{\url{fr.vikidia.org}} and Wikimini\footnote{\url{fr.wikimini.org}} for children, and Wikipedia article on difficult topics regarding the adult age range. Newspaper texts come from various newspapers and magazines like \textit{Le P'titLibé}\footnote{\url{https://ptitlibe.liberation.fr/}} or \textit{Albert}\footnote{\url{https://www.journal-albert.fr/}} for children, whereas they come from usual newspapers for adults (\textit{L'Humanité}, \textit{Le Monde diplomatique}, etc.). Finally, fictional texts come form stories or novels with various authors.

The corpus is balanced in terms of text genres and age ranges. However, no specific attention has been paid on balancing the topics across the age ranges and genres. This is simply because multiplying the constraints makes it impossible to build a corpus large enough to perform machine learning\footnote{In this regard, the corpus could have been much large if text genres were not balanced since encyclopedic data for children is easy to massively collect.}.

Text lengths are very different according to their genre and target age, especially regarding novels which can be very long for adults. The information about length could be taken by the models as a clue in favor of high ages. To avoid this bias, all texts larger than $10,000$ characters have been segmented on paragraph boundaries\footnote{In practice, these boundaries are detected by blank lines instead of simply line returns since the latter strategy would badly segment dialogues.} with a segment length of $5,000$ characters approximately. This segmentation avoids the biases due to the genre and target age too. Moreover, it is consistent with one of our goals to index online texts (which are rarely presented as one very long page but rather paginated) or to assist writers (they may probably not want to study their whole text but rather portions of it).

The whole dataset is partitioned into train/validation/test sets with respective proportions of $68.3 / 16.5 / 15.2\%$ in terms of texts, and $66.7/16.9/16.5\%$ regarding the sentences. All texts segments coming from a same original text are gathered in the same subset to guarantee that the validation and test data are completely unseen data after training.
In all subsets, fictions count more texts and sentences than two other genres of encyclopedia and newspaper. 
Moreover, the average age ranges and mean ages per subset and per genre vary slightly. Especially, it can be noticed that the ages from the validation set differ more to the training set than the test set does.

\subsection{Statistics on Ages}



\begin{figure*}[t]
\begin{center}
\includegraphics[scale=0.8]{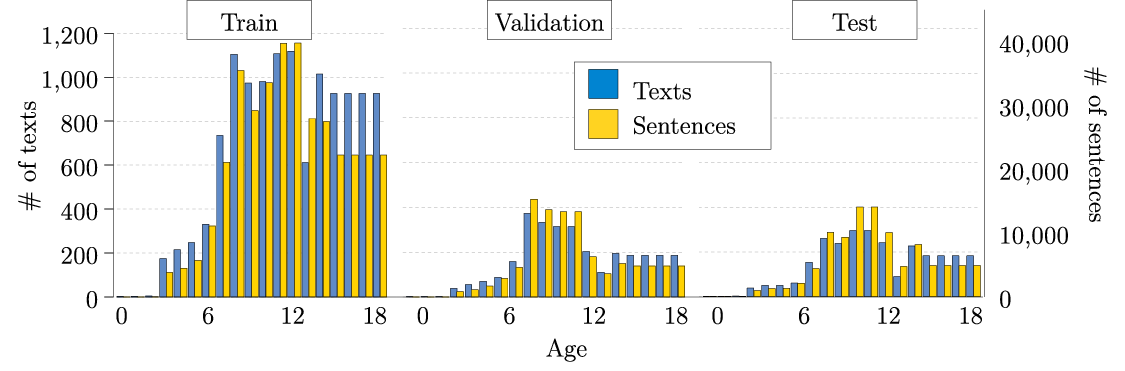}
\caption{\label{fig:data_distribution}Distribution of the texts and sentences over the different ages in the train, validation, and test sets}
\end{center}
\vspace{-3mm}
\end{figure*}

The distributions of the texts and sentences per age are shown in Figure~\ref{fig:data_distribution}. One text or sentence with age range $[a, b]$ is counted up for an age $x$ as soon as $x \in [a, b]$. This figure shows that the ages from 8 to 12 years are slightly more represented, especially regarding sentences. This corresponds to the period where children can read but are still considered as children (as opposed to teenagers).
In contrast, few texts/sentences correspond to the ages between 0 to 3 years. This is normal as the children of these ages are too young to read and their attention time is smaller. Hence, these texts are supposed to be narrated, and they are short.
%

The detailed statistics of the dataset per age range are shown in Table~\ref{table:dataStat_ageRange}. The dataset contains texts for 32 different age ranges. However, the train, validation and test sets individually does not cover the same age ranges. The three sets contain 29, 19, and 24 age ranges, respectively where 18 age ranges are in common.

\begin{table*}[ht!]
\centering
	\renewcommand{\arraystretch}{1.3}
\small
	\begin{tabular}{|c|c||r|r|r||r|r|r||r|r|r|}
	 \hline
     \multicolumn{2}{|c|}{\textbf{Age}} & \multicolumn{3}{c|}{\textbf{Train}} & \multicolumn{3}{c|}{\textbf{Validation}} & \multicolumn{3}{c|}{\textbf{Test}}  \\
      \hline
	  \textbf{Range} & \textbf{Mean}  & \textbf{Texts}  & \textbf{Sent.}  & \textbf{Tokens}  & \textbf{Texts}  & \textbf{Sent.}  & \textbf{Tokens}  & \textbf{Texts}  & \textbf{Sent.}  & \textbf{Tokens} \\ \hline
\range{0}{2} & 1.0 & 1 & 6 & 73 & -- & -- & -- & -- & -- & -- \\
\range{0}{3} & 1.5 & 3 & 29 & 255 & 2 & 33 & 277 & 2 & 60 & 686 \\
\range{2}{4} & 3.0 & -- & -- & -- & -- & -- & -- & 1 & 17 & 258 \\
\range{2}{7} & 4.5 & 1 & 20 & 486 & -- & -- & -- & -- & -- & -- \\
\range{3}{5} & 4.0 & 28 & 1,116 & 13,318 & 7 & 241 & 3,029 & 8 & 206 & 2,402 \\
\range{3}{6} & 4.5 & 79 & 1,552 & 25,660 & 11 & 197 & 3,215 & 16 & 360 & 5,143 \\
\range{3}{7} & 5.0 & 15 & 321 & 5,034 & 7 & 104 & 1,669 & 4 & 88 & 1,225 \\
\range{3}{8} & 5.5 & 29 & 683 & 12,761 & 11 & 262 & 4,371 & 5 & 107 & 1,925 \\
\range{3}{9} & 6.0 & 20 & 459 & 7,918 & 1 & 23 & 356 & 3 & 132 & 2,229 \\
\range{4}{6} & 5.0 & -- & -- & -- & 1 & 58 & 650 & 1 & 121 & 1,121 \\
\range{4}{7} & 5.5 & 1 & 20 & 336 & -- & -- & -- & 1 & 37 & 948 \\
\range{4}{8} & 6.0 & 31 & 494 & 6,917 & 14 & 219 & 3,318 & 10 & 166 & 2,393 \\
\range{4}{9} & 6.5 & 11 & 217 & 4,614 & 3 & 69 & 1,358 & 2 & 59 & 1,254 \\
\range{5}{7} & 6.0 & 5 & 275 & 4,301 & -- & -- & -- & -- & -- & -- \\
\range{5}{8} & 6.5 & 1 & 38 & 623 & -- & -- & -- & -- & -- & -- \\
\range{5}{9} & 7.0 & 25 & 989 & 17,420 & 15 & 592 & 11,493 & 1 & 48 & 948 \\
\range{5}{12} & 8.5 & 1 & 27 & 632 & -- & -- & -- & -- & -- & -- \\
\range{6}{8} & 7.0 & 70 & 5,571 & 69,513 & 16 & 1,208 & 15,494 & 7 & 479 & 6,311 \\
\range{6}{9} & 7.5 & 42 & 1,365 & 24,396 & 8 & 276 & 5,414 & 8 & 361 & 7,219 \\
\range{6}{12} & 9.0 & -- & -- & -- & -- & -- & -- & 3 & 167 & 2,879 \\
\range{7}{11} & 9.0 & 17 & 1,096 & 13,608 & -- & -- & -- & 4 & 279 & 3,306 \\
\range{7}{12} & 9.5 & 466 & 11,231 & 232,898 & 83 & 1,989 & 41,918 & 106 & 2,477 & 51,352 \\
\range{8}{10} & 9.0 & 57 & 4,915 & 58,352 & -- & -- & -- & -- & -- & -- \\
\range{8}{11} & 9.5 & 29 & 1,706 & 27,799 & 120 & 7,562 & 104,882 & 62 & 4,232 & 56,106 \\
\range{8}{12} & 10.0 & 28 & 765 & 13,507 & 2 & 80 & 1,390 & 4 & 129 & 2,283 \\
\range{8}{13} & 10.5 & 277 & 8,781 & 223,954 & 101 & 3,356 & 79,749 & 46 & 1,456 & 36,687 \\
\range{10}{12} & 11.0 & 12 & 751 & 9,707 & 8 & 645 & 7,105 & 38 & 2,533 & 32,724 \\
\range{10}{13} & 11.5 & 63 & 4,169 & 57,710 & -- & -- & -- & -- & -- & -- \\
\range{10}{14} & 12.0 & 30 & 2,815 & 28,214 & -- & -- & -- & 32 & 2,790 & 27,946 \\
\range{11}{13} & 12.0 & 183 & 11,593 & 175,219 & -- & -- & -- & -- & -- & -- \\
\range{12}{14} & 13.0 & 58 & 2,844 & 43,870 & 9 & 365 & 6,243 & 12 & 503 & 8,401 \\
\range{14}{18} & 16.0 & 926 & 24,034 & 654,148 & 186 & 4,982 & 129,408 & 183 & 4,894 & 125,796 \\
\hline
\multicolumn{2}{|c|}{\textbf{Total}} & \textbf{2,509} & \textbf{87,882} & \textbf{1,733,243} & \textbf{605} & \textbf{22,261} & \textbf{421,339} & \textbf{559} & \textbf{21,701} & \textbf{381,542} \\
\hline
    \end{tabular}
\caption{\label{table:dataStat_ageRange} Number of texts, sentences and tokens for each age range in the train, validation and test sets. Age ranges are sorted according to their lower bound first. The sign -- is used when no data is available.}
\end{table*}




%% file: section_5_metrics.tex
\section{Evaluation metrics}
\label{sec:metrics}
\newcommand{\twodvector}[2]{\left(\begin{array}{@{}c@{}}#1\\#2\end{array}\right)}

This section presents and studies different metrics to compute the error between a reference interval $r = [a, b]$ provided by the annotations, and a hypothesis interval $h = [c, d]$. While this study does not pretend to be exhaustive, the objective is to highlight the main questions for the age recommendation task and to come up with one or few metrics for the remainder of the paper, hopefully future work in the community as well.

Overall, most approaches vary according to the meaning behind the notion of interval. On the one hand, one can assume that an interval $[x, y]$ is exact, that is all children whose age is in this interval will understand the text. Besides, uncertainty can be taken into account to reflect the fact that the real (reference) range/value is unknown and  the only available information is that this is somewhere within $[x, y]$. As previously discussed, this uncertainty assumption seems particularly reasonable for our data since their underlying annotation process by the authors and editors is not perfectly understood.

Below, the section presents several approaches and their most relevant metrics\footnote{Several other variants have been studied but are not reported here since they do not bring interesting theoretical or experimental results.}. Finally, these metrics are compared and discussed on a study case.

\subsection{Usual vector metrics}
\label{sec:vector_metrics}

Considering ranges as 2D vectors $\vec{r} = \twodvector{a}{b}$ and $\vec{h} = \twodvector{c}{d}$, usual metrics are the $L_1$ and $L_2$ distances (``absolute Error'' and ``root squared error''), that is:
 \begin{equation}
  L_1(\vec{h} - \vec{r}) = \mid c - a\mid + \mid d - b\mid~,
    \vspace{-1mm}
 \end{equation}
 \begin{equation}
  \mbox{and~~} L_2(\vec{h} - \vec{r}) = \sqrt{(c - a)^2 + (d - b)^2}~.
  \vspace{-1mm}
 \end{equation}


Intuitively, one could argue that it is better when the hypothesis interval is included in the reference one, rather than the contrary. This can taken into consideration by measuring the angle $\theta$ between the vector $\vec{h}-\vec{r} = \twodvector{c - a}{d - b}$ and $\vec{\imath} = \twodvector{1}{-1}$\footnote{An illustration of this principle is provided in Appendix~\ref{app:metrics}.}.
Mathematically, this is computed as:
\begin{equation}
 \cos(\theta) = \cos\,(~\vec{h}-\vec{r}~,~\,\vec{\imath}~)
              = \left\{\begin{array}{cl}
 1 & \mbox{if } \vec{r} = \vec{h} \mbox{ ,} \\
 \frac{c - a - d + b}{\sqrt{2} \, L_2(\vec{h}-\vec{r})} & \mbox{otherwise~.} \\
 \end{array}\right.
   \vspace{-1mm}
\end{equation}
$L_2$ and $\cos(\theta)$ can then be linearly interpolated as follows to provide a new metric:
\begin{equation}
 \theta\mbox{-}L_2(\vec{r}, \vec{h}) = L_2(\vec{h} - \vec{r})+ \alpha (1 - \cos(\theta))~,
 \vspace{-1mm}
\end{equation}
where $\alpha$ is an empirically tunable weight, and $1 - \cos(\theta)$ ranges in $[0, 2]$.


The main drawback of all these metrics is that the lower and upper bounds of the intervals are processed independently, whereas it is clear that they are not independent.

\subsection{Jaccard distance}
\label{sec:jaccard_metrics}

Jaccard distance, which is usually applied on sets, can be extended to intervals and can be used to evaluate age ranges. The definition is as follows:
\begin{equation}
    J([a,b], [c, d]) = 1 - \frac{\|~[a, b] \cap [c ,d]~\|}{\|~[a, b] \cup [c ,d]~\|}~,
    \vspace{-1mm}
\end{equation}
where $\|.\|$ denotes the norm of an interval.
The resulting measure ranges in $[0, 1]$ where $0$ means that the 2 intervals do not overlap at all, whereas $1$ is returned if they are equal.

The Jaccard distance is debatable for various reasons. First, it does not consider the size of the gap when the intervals do not overlap. Second, $J$ is symmetric, whereas the intuition would argue for a prevalence of the reference interval $r$ over the hypothesized one $h$. Finally, $J(r, h)$ normalizes the error according to the size of the intervals. Hence, big errors on large intervals are equivalent to small errors on small intervals. This last argument can be overwhelmed by scaling $J$ according the size of the intervals, for instance, using the average interval size:
\begin{equation}
 J_{\mbox{year}}([a,b], [c, d]) = \frac{\mid b-a\mid+\mid d-c\mid}{2}\times J([a,b], [c, d])~.
 \vspace{-1mm}
\end{equation}

\subsection{Mean-based metrics}
\label{sec:mean_metrics}

Assuming that the bounds of a given age range are somewhat uncertain, one may argue that the interval should be reduced to its center/mean value.
%
When applying this principle on both intervals, this defines a mean-to-mean error, denoted as follows:
\begin{equation}
    \mu{}E_{a, b}(c, d) = \mid \mu(a, b) - \mu(c, d) \mid~,
    \vspace{-1mm}
\end{equation}
where $\mu(x, y)$ is the mean of $x$ and $y$.
While this measure may look oversimplistic, we believe that it is interesting because it is easily interpretable.

This metric can be relaxed further by considering that a recommendation is acceptable (i.e., no error) as soon as the mean of the hypothesis interval lies in the reference interval. This is defined as the Bound Error (BE):
 \begin{equation}
    BE_{a, b}(c, d) = \left\{
    \begin{array}{l@{~~~~}l}
        0               & \mbox{if }\mu(c, d) \in [a ,b]~, \\
        a - \mu(c, d) & \mbox{if }\mu(c, d) < a~, \\
        \mu(c, d) - b & \mbox{if }\mu(c, d) > b~.
    \end{array}
    \right.
    \vspace{-1mm}
\end{equation}

%
%
%
%
%
%
%

\subsection{Integral of local errors}
\label{sec:local_error_metrics}

In a last approach, the error between two age ranges can be seen as the sum of a local errors $e(x)$ for age $x$ over the recommendation domain $[0, 18]$. In practice, this domain can be reduced to the smallest segment which both includes the reference and hypothesis intervals because there is no error to be counted outside of it. Hence, we define the Integral Error (IE) as:
\begin{equation}
    IE_{a,b}(c,d) = \int_{\min(a, c)}^{\max(b, d)} e_{a,b,c,d}(x) \, dx~.
    \vspace{-1mm}
\end{equation}
IE is interesting since it enables a large panel of logics through the definition of different functions $e$.
\begin{figure}[t!]
 \centering
 \includegraphics[scale=0.8]{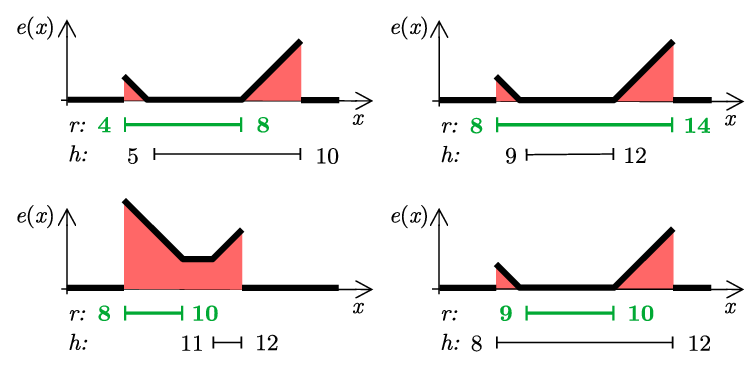}
 \caption{\label{fig:range_metrics_integrals}Evolution of the local error $e$ and the resulting global error IE (filled areas below bold lines) for prototypical situations of the reference ($r$) and hypothesis ($h$) ranges.}
\end{figure}

A straightforward definition for $e(x)$ consists in measuring the distance btween $x$ and $r$, the distance between $x$ and $h$, and to sum the two distances. The distance between $x$ and a given interval $[s, t]$ is null if $x \in [s, t]$, and the absolute distance to the closest bound otherwise. This can be written as follows:
\begin{equation}
    e_{a,b,c,d}(x) = \left\{ \begin{array}{ll}
        0                      & \mbox{if } x \in [a, b] \mbox{ and } x \in [c, d] \mbox{~,} \\[2mm]
        \min(\mid x - c\mid, \mid x - d\mid) & \mbox{if } x \in [a, b] \mbox{ and } x \notin [c, d] \mbox{~,} \\[2mm]
        \min(\mid x - a\mid, \mid x - b\mid) & \mbox{if } x \notin [a, b] \mbox{ and } x \in [c, d] \mbox{~,} \\[2mm]
        \multicolumn{2}{l}{\min(\mid x - c\mid, \mid x - d\mid) + \min(\mid x - a\mid, \mid x - b\mid)} \\
        & \mbox{if } x \notin [a, b] \mbox{ and } x \notin [c, d] \mbox{~.} \\
    \end{array} \right.
    \vspace{-1mm}
\end{equation}
After integrating and applying square root to express the result as ``years'' (instead of ``squared~years''), this brings to the following Symmetric Integral Error (Sym-IE):
\begin{equation}
Sym\mbox{-}IE_{a,b}(c, d) = \sqrt{\frac{(a - c)^2 + (d - b)^2}{2}} = \frac{L_2(\vec{h}-\vec{r})}{\sqrt{2}}~.
\vspace{-1mm}
\end{equation}
Figure~\ref{fig:range_metrics_integrals} illustrates the behavior of $e(x)$ (solid line) and $Sym\mbox{-}IE$ (hatched area) on different cases.




The frame of local errors can be easily extended. For instance, Sym-IE can be generalized by weighting differently the local error due to $r$ and due to $h$ with a factor $\beta \in [0, 1]$:
\begin{equation}
    e_{a,b,c,d}(x) = \left\{ \begin{array}{ll}
        0                      & \mbox{if } x \in [a, b] \mbox{ and } x \in [c, d] \mbox{~,} \\[2mm]
        \beta  \min(\mid x - c\mid, \mid x - d\mid) & \mbox{if } x \in [a, b] \mbox{ and } x \notin [c, d] \mbox{~,} \\[2mm]
        (1 - \beta)  \min(\mid x - a\mid, \mid x - b\mid) & \mbox{if } x \notin [a, b] \mbox{ and } x \in [c, d] \mbox{~,} \\[2mm]
        \multicolumn{2}{l}{\beta  \min(\mid x - c\mid, \mid x - d\mid) + (1 - \beta)  \min(\mid x - a\mid, \mid x - b\mid)} \\
        & \mbox{if } x \notin [a, b] \mbox{ and } x \notin [c, d] \mbox{~,} \\
    \end{array} \right.
    \vspace{-1mm}
\end{equation}
A global error, called $\beta$-IE, is then defined as:
%
%
\begin{equation}
 \beta\mbox{-}IE_{a,b}(c, d) =  \sqrt{
 \begin{array}{l}
\beta \times (\max(0, c-a)^2 + \max(0, b-d)^2) \\
                                + (1-\beta) \times (\max(0, a-c)^2 + \max(0, d-b)^2)
                                \end{array}}
                                ~.
    \vspace{-1mm}
\end{equation}
The setting $\beta = 0.5$ corresponds to Sym-IE, while $\beta < 0.5$ gives more importance to fitting the reference interval than the hypothesis.

Other extensions could be imagined, which we believe makes this approach better than the previous ones on a theoretical point of view. For instance, one could integrate the fact that errors on early ages are more impacting than those on higher ages, since the cognitive development of children slows down as they grow up.

\subsection{Study case and discussion}
\label{sec:metrics_experiments}
Among all the previously defined metrics, a good metric would behave in the same way as a human would intuitively do. This means that, given a reference age range, a good metric would rank a set of hypothesis ranges in the same order as a human.
To do so, we designed a set of $n = 20$ hypothesis intervals ($h_i$, $1 \leq i \leq n$) with respect to a reference $r = [8, 12]$, and ranked them in an oracle way. These hypotheses reflect typical situations that one would want the metric to distinguish appropriately.
Given a metric $d$, all hypotheses $h_i$ can sorted in the ascending order considering $d(r, h_i)$, and associated to their rank $\rho_d(i)$. Then, the quality of $d$ is measured by comparing the ranking $\rho_d$ with the oracle ranking $\rho^*$ using the average Spearman's footrule distance $S$ with the oracle ranking~\cite{spearman1904proof} (the lower the better):
\begin{equation}
 S(\rho^*, \rho_d) = \frac{1}{n} \sum_{i=1}^{n} \mid\rho^*(i) - \rho_d(i)\mid~.
 \vspace{-1mm}
\end{equation}

Table~\ref{tab:range_metrics_ranks} shows the rankings for each metric\footnote{The corresponding values for $d(r, h_i)$ are provided in Appendix~\ref{app:metrics}.} along with the resulting value for $S$. The metrics under investigation are compared to what would do a random metric (i.e., $d(r, h_i)$ is a random positive real number)\footnote{Let one note that the expected Spearman's footrule distance $S$ for this random metric is an empirical approximation since no analytical form for the expectation of Spearman's footrule distance $S$ is known in the litterature.}. Values of the parameters $\alpha = 0.5$ and $\beta = \frac{1}{3}$ for the parametric metrics are those leading to the best results for $S$.\footnote{Eventhough this would be a methodological issue for some experiments, we judged that this optimization process is acceptable since the objective is not to provide the best parameters but to understand the way that the metrics behave.}

\begin{table*}[t!]
\centering
 \includegraphics[scale=0.8]{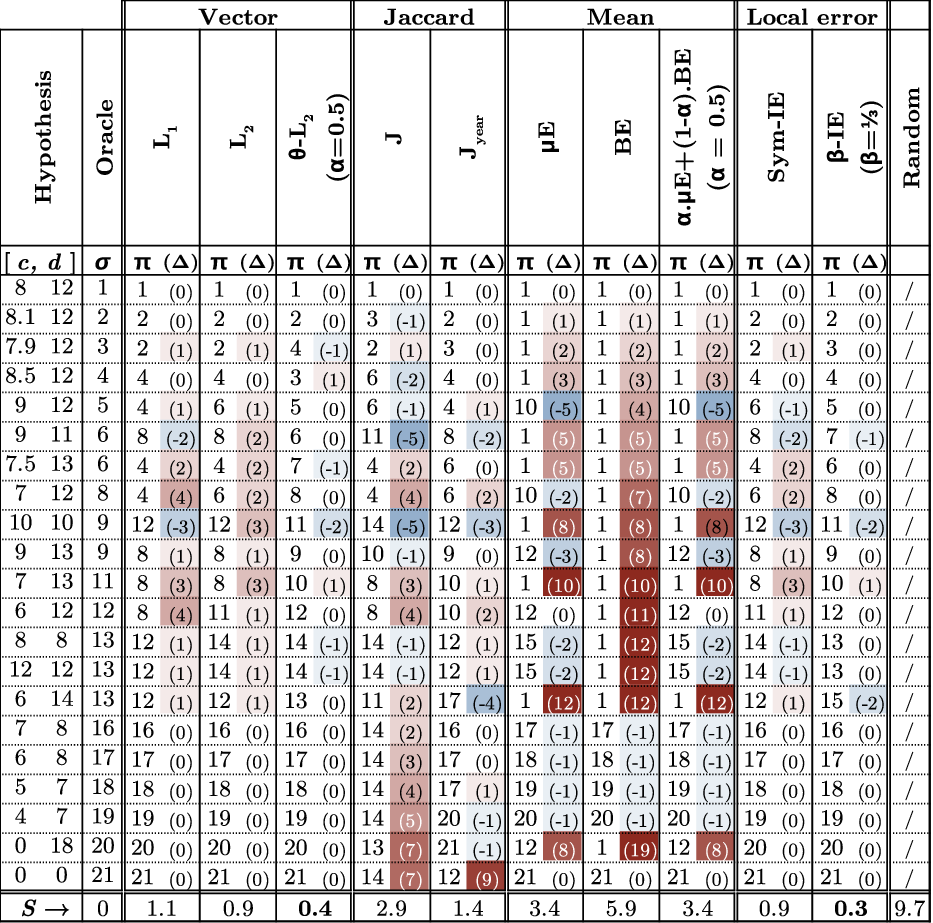}
\caption{\label{tab:range_metrics_ranks} Ranks given by the different metrics for various hypothesis intervals $[c, d]$ according to the reference interval $[a, b] = [8, 12]$. The rank difference with the Oracle is given in parenthesis ($\Delta)$, and Spearman's footrule distances $S$ against the Oracle are given by the last row.}

\end{table*}

Overall, it appears that the values for $S$ are very varied. Some metrics lead to high values. For $J$, this is consistent with the drawbacks previously highlighted. For $BE$, it appears that this comes from the fact that many hypotheses cannot be (purposely) distinguished. As shown in the tables, a linear interpolation with $\mu$E can easily alleviate, even for very small values of $\alpha$. The same phenomenon is observed with Asym-IE. At the opposite, the best metrics reach very low values, which denote a strong correlation with the human judgement ($\beta$-IE and $\theta$-L$_2$).

Even if the hypothesis ranges used in this experiment do not perfectly fit the expected distribution of the hypotheses on the real data, we believe that this study provides a valuable basis to rank the models experimented in this paper.
As such, the results will be provided in terms of $\beta$-IE$_{\beta = 0.4}$, $\theta$-$L_{2, \beta = 0.4}$. Additionnally, the results will be reported in terms of $\mu$E (error between the means) because this is easily interpretable and corresponds to the situation where only one number is expected from age recommendation, instead of a range.

%% file: section_6_prediction_experiments_new.tex
\section{Age Recommendation Experiments}
\label{sec:age_pred_experiments}
We considers age recommendation (prediction of a target age) as a regression task. This section details the different approaches and machine learning models for age prediction from a text/sentence, and shows the experimental results. 
\subsection{Experiment Protocols}
The age recommendation approaches studied here are (a) OneShot text-level (by considering a text as a whole), (b) sentence-level, and (c) aggregation text-level. In aggregation approach, the sentence-level recommendations are aggregated to make a text-level global recommendation.
Moreover, in each approach, a text/sentence is represented with a vector of several features (as real numbers). This paper studies mainly two kinds of features: word embedding features and linguistic expert features.
For the embedding features, we use two pretrained embeddings, FastText and CamemBERT. FastText is a CBOW-based word embeddings (dimension of $300$) which is pre-trained on French Wikipedia using fastText~\citep{grave2018learning}. Unlike FastText, CamemBERT~\citep{martin-etal-2020-camembert} is a BERT-based contextual embeddings (dimension of $768$), trained on the French subcorpus of a multilingual corpus named OSCAR. However, the linguistic expert features are a set of $107$ features of $9$ categories adapted from the literature of related studies. Table~\ref{table:expert_features_cat} presents the $9$ categories with their feature counts\footnote{ Appendix~\ref{app:expert_features} describes the feature categories in detail.}.
%
\begin{table*}[t!]
\centering
\small
	\renewcommand{\arraystretch}{1.3}
\begin{tabular}{|l|l|}
    \hline
    \textbf{Feature Category} & \textbf{\# Features} \\ \hline
    Lexicon & 5 \\
    Graphemes & 6 \\
    Morphosyntax & 7 \\
    Verbal Tenses & 24 \\
    Person/Number & 5 \\
    Syntactic Dependencies & 8 \\
    Logical Connectors & 16 \\
    Phonetics & 9 \\
    Sentiments & 27 \\
    \hline
    \textbf{Overall} & \textbf{107} \\
    \hline
\end{tabular}
	\caption{\label{table:expert_features_cat} Linguistic expert features of 9 categories inherited from the related study}
\end{table*}


%
With the word embeddings and linguistic expert features, different representations of a sentence and a text are detailed in Table~\ref{table:sentence_and_text_representations}.
Several regression models are trained with these representations of features to predict the target age from a text/sentence. Here we study the following regression models for age recommendation.

\begin{table*}[t!]
\centering
\small
	\renewcommand{\arraystretch}{1.4}
\begin{tabular}{|l|l|l|}
    \hline
    \textbf{Data} & \textbf{Representation with Features} & \textbf{Expression} \\ \hline
    \multirow{3}{*}{Sentence} & 1. Avg. of token/word-vectors from FastText & $S_{FT}=\overline{V_{token}(FastText)}$ \\ 
    & 2. Avg. of token/word-vectors from CamemBERT & $S_{CB}=\overline{V_{token}(CamemBERT)}$ \\ 
    & 3. Sentence-level linguistic expert features & $S_{Expert}$\\
    \hline
    \multirow{3}{*}{Text} & 1. Avg. of sentence-vectors from FastText & $T_{FT}=\overline{S_{FT}}$ \\ 
    & 2. Sequence of sentence-vectors from CamemBERT & $T_{CB}=\{S_{CB}^1,S_{CB}^2,...,S_{CB}^N\}$ \\ 
    & 3. Avg. of sentence-level linguistic expert features & $T_{Expert}=\overline{S_{Expert}}$\\
    \hline
\end{tabular}
\caption{
\label{table:sentence_and_text_representations} Different representations of a sentence and a text to train an age-recommendation model}
\end{table*}

\subsubsection{Regression Models}
\label{sec:models}
\textbf{CamemBERT}, a French language model relies on a RoBERTa architecture~\citep{DBLP:journals/corr/abs-1907-11692}, which includes positional encoding and multi-head masking with attention mechanism to learn the language representation. It bidirectionally learns contextual relations from words in all positions unlike the sequential learning in RNN architecture. A pretrained CamemBERT provides also the contextual vector of a word or sentence (mean of the word-vectors in a sentence). CamemBERT offers performing single-output regression task. For age-range recommendation, given a sentence as input, two CamemBERT models are employed to recommend the lower bound and upper bound separately.
\par
\smallskip
\noindent A \textbf{RNN} model e.g., GRU~\citep{cho2014learning} are able to learn one-way long-term dependencies of a sequence. Such a model is highly relevant to text-based recommendations as a text holds the necessary properties of sequence. 
Although CamemBERT allows regression on long texts (maximum 512 tokens), many texts in our dataset are much larger than this limit. Therefore, for OneShot text-level age recommendation, this paper uses a GRU network fed with $T_{CB}$. We studied RNN models also for the sentence-level age recommendation but the results were not so good to report here. 
\par
\smallskip
\noindent A \textbf{Feed-forward} is a multilayer perceptron network with an input layer, hidden layer(s), and an output layer. With less parameters, it is comparatively a lighter deep learning architecture than the advanced ones e.g., BERT, transformers etc. In age recommendation task, the Feed-forward model is fed with the feature-vectors of size N representing a text/sentence, and directly outputs the two bounds of an age-range.
\par
\smallskip
\noindent A \textbf{Random Forest} is an ensemble learning model to perform classification or regression task which combines multiple decision trees to make the final prediction. A Random Forest reduces the overfitting and gets rid of the limitations of a decision tree algorithm. It offers explainability of the recommendations. We employ Random Forest regressor to directly predict the lower bound and upper bound of an age-range like the RNN and Feed-forward ones.
\par
\begin{figure}[t!]
\begin{center}
\includegraphics[width=0.75\linewidth]{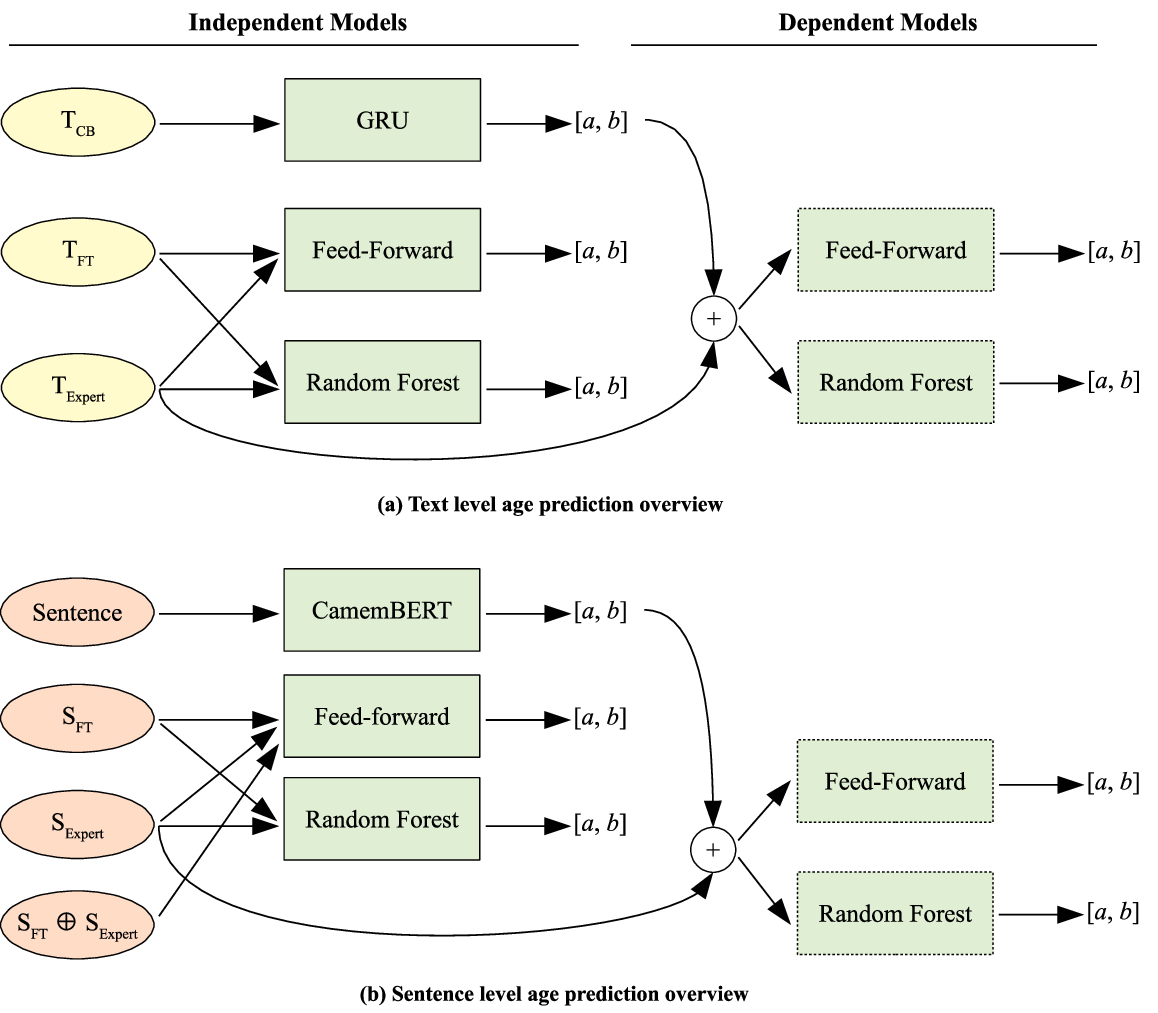}
\caption{\label{fig:t2k_features_n_models} Overview of the text-level and sentence-level age recommendation models}
\end{center}
\end{figure}

\smallskip
\noindent \textbf{Combinations of the different features and regression models} 
The regression models are fed with different types of features regarding either age recommendation from text or sentence. Figure~\ref{fig:t2k_features_n_models} shows the overview of different regression models and their associated inputs. 
A model that takes embedding features and expert features as input is named as an \textit{independent model}. In contrast, we name a model as \textit{dependent model} when the model's input is the output of another model (in particular, GRU and CamemBERT), concatenated with the expert-features. Thus Feed-forward and Random Forest act as both independent and dependent models based on the input. However, the GRU (for text-level recommendation) and CamemBERT (for sentence-level recommendation) models are used only as the independent models.
\par
\begin{figure}[t!]
\begin{center}
\includegraphics[width=0.75\linewidth]{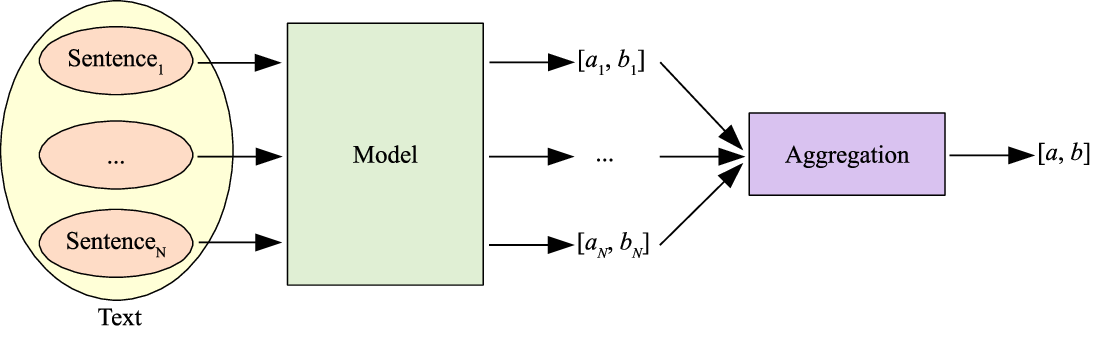}
\caption{\label{fig:prediction_aggregation}Text-level age recommendation by aggregating sentence-level recommendations}
\end{center}
\end{figure}
\begin{equation}
grade \: level = 0.39 \left(\frac{total \: words}{total \: sentences}\right) + 11.8 \left(\frac{total \: syllables}{total \: words}\right) - 15.59
\label{eq:flesh-kincaid}
\vspace{10mm}
\end{equation}
\smallskip
\noindent \textbf{Sentence aggregation, Grade-level, and Naive approach}
We also study aggregation models to recommend the age for a text from its sentence level recommendations as shown in Figure~\ref{fig:prediction_aggregation}. This study explores different aggregation approaches e.g., mean, median, and RNN but finds mean aggregation as the most robust and reliable. Therefore, this paper reports only the mean aggregation, $\left[a, \: b\right] = \frac{\sum_{i=1}^{N} \left[a_i, \: b_i\right]}{N}$, where $\left[a_i, \: b_i\right]$s are recommended ages at the sentence level, and $N$ is the number of sentences in a text.
This paper further uses \textit{Flesch-Kincaid grade-level score}~\citep{FleshKincaidGradeScore} which computes a grade-level (Eq.~\ref{eq:flesh-kincaid}) for an input text. It is a very simple model but widely studied baseline in text-readability analysis. For a text, to derive the recommended age from Flesch-Kincaid score, we use a base-age of $5.5$, and add it to the grade-level. Thus, for example, regarding a text, if the grade-level is $4.5$, the recommended age becomes $4.5 + 5.5 = 10$ years. As Flesch-Kincaid score returns a single grade-level, this paper considers the derived recommended age as a mean-age ($\mu$), and it is comparable with the recommendations of other models by $\mu\mbox{-}score$ only. Finally we use a \textit{Naive} model, for any text or sentence as input, it outputs the mean values of age-ranges ($[9.86, \: 13.78]$ for text-level and $[9.73, \: 13.26]$ for sentence-level) computed from the training dataset. This naive approach helps us to justify the other models, either they learn well from the data or not.

\subsubsection{Training and Parameter Tuning}
The age recommendation models are built on the train dataset. Further, to confirm the stability, each model is fine-tuned with several hyper-parameters on the validation dataset.
Table~\ref{table:hyper_param} details the different hyper-parameters calibrated in the models.

\begin{table*}[t!]
\centering
\small
	\renewcommand{\arraystretch}{1.6}
 \begin{tabular}{|l|l|l|}
	 \hline
\textbf{Model} & \textbf{Text Level Prediction} & \textbf{Sentence Level Prediction} \\ \hline
\textbf{Feed-forward} & 
\makecell[l]{hidden\_layers = 6, hidden\_units = 200, \\ activation = ReLU, epoch = 500} & \makecell[l]{hidden\_layers = 6, hidden\_units = 200, \\ activation = ReLU, epoch = 500} \\ \hline
\textbf{Random Forest} & \makecell[l]{nb\_estimators = 100, criterion = 'gini', \\ nb\_leaf\_node = $\infty$} & \makecell[l]{nb\_estimators = 100, criterion = 'gini', \\ nb\_leaf\_node = $\infty$}\\ \hline
\textbf{GRU} & \makecell[l]{units = 128,\\ activation = tanh, batch\_size = 128,\\ drop\_out = 0.20, epoch = 50} & Not Used \\ \hline
\textbf{CamemBERT} & Not Used & \makecell[l]{max\_sentence\_len = 100, lr = 1e-5,\\ batch\_size = 32, epoch = 3} \\
\hline
\end{tabular}
\caption{\label{table:hyper_param} Hyper-parameters used in different models for fine-tuning purpose}
\end{table*}

\subsection{Age Recommendation Scores}
Here we present the scores obtained from the different age recommendation models. Basically these scores refer to the errors made by the different recommendation models. To confirm the stability of the age recommendation models, this section reports the $\mu\mbox{-}score$ of each model on both validation and test datasets. However, the $\theta\mbox{-}L_2$ and $\beta\mbox{-}IE$ scores of each model are presented for test dataset only.

\subsubsection{OneShot text-level recommendation}
In OneShot text-level age recommendation, we train GRU, Feed-forward and Random Forest models with the different types of features as shown in Fig.~\ref{fig:t2k_features_n_models}(a).
Table~\ref{table:OneShot text level muE} shows the $\mu{}E$ scores of different models on the validation and test datasets. All the models except Flesch-Kincaid significantly outperform the Naive approach. Flesch-Kincaid method gets the worst $\mu{}E$ score and it is not very surprising as this method is very simple. The Naive approach obtains better score than Flesch-Kincaid method because the age-range of a large portion of the dataset is approximately same as the Naive approach computes.
However, as an independent model, GRU with $T_{CB}$ features results the best scores for the three genres Encyclopedia (E), Newspaper (N) and Fiction (F) on both validation and test datasets. With $T_{FT}$ features, the Feed-forward model performs the second best.
Moreover, Random Forest performs better than the Feed-forward model when both are fed with $T_{Expert}$ features. Interestingly, as a dependent model, Random Forest with the combination of GRU recommended min\_age, max\_age, and $T_{Expert}$ features obtains a little better score than the GRU model. Also in terms of the proposed metrics, $\theta\mbox{-}L_2$ and  $\beta\mbox{-}IE$, GRU with $T_{CB}$ features outperforms other models as shown in Table \ref{table:OneShot text-level proposed metrics}. These results signify the effectiveness of $T_{CB}$ for OneShot text-level age recommendation. It is as expected because CamemBERT holds the contextual information of the words in a sentence, and such information helps a neural network model to extract useful hidden features.

\begin{table*}[t!]
\centering
\small
	\renewcommand{\arraystretch}{1.3}
	
 \begin{tabular}{|l|l|l|l|l||l|l|l|l|}
	 \hline
     & \multicolumn{4}{c|}{\textbf{Validation}} & \multicolumn{4}{c|}{\textbf{Test}}\\
      \hline
	  \textbf{Model/Features} & \textbf{E} & \textbf{N} & \textbf{F} & \textbf{All} & \textbf{E} & \textbf{N} & \textbf{F} & \textbf{All} \\ \hline
Naive & 2.80 & 2.94 & 3.78 & 3.25 & 3.24 & 2.94 & 3.12 & 3.10 \\
Flesch-Kincaid & 6.22 & 4.97 & 5.88 & 5.76 & 6.92 & 4.61 & 4.78 & 5.23 \\
GRU/T\textsubscript{CB} & 1.42 & 0.30 & 1.97 & \textbf{1.39} & 1.18 & 0.35 & 1.40 & \textbf{0.98} \\
Feed-forward/T\textsubscript{FT} & 1.70 & 1.10 & 2.61 & 1.96 & 1.72 & 1.01 & 1.79 & 1.52 \\
Feed-forward/T\textsubscript{Expert} & 2.31 & 1.95 & 2.71 & 2.42 & 2.15 & 1.90 & 2.51 & 2.22 \\
Random Forest/T\textsubscript{FT} & 2.08 & 1.33 & 2.84 & 2.23 & 1.80 & 1.21 & 2.19 & 1.78 \\
Random Forest/T\textsubscript{Expert} & 2.04 & 1.40 & 2.42 & 2.06 & 1.70 & 1.35 & 2.18 & 1.78  \\ \hline
Feed-forward/(GRU\_pred.+T\textsubscript{Expert})* & 1.72 & 0.79 & 2.26 & 1.72 & 1.42 & 0.85 & 1.70 & 1.34 \\
Random Forest/(GRU\_pred.+T\textsubscript{Expert})* & 1.33 & 0.23 & 1.95 & \textbf{1.34} & 1.05 & 0.26 & 1.37 & \textbf{0.91} \\
\hline
    \end{tabular}
\caption{\label{table:OneShot text level muE} OneShot text-level $\mu{}E$ scores (* indicates dependent model)}
\end{table*}
%
\begin{table*}[t!]
\centering
\small
	\renewcommand{\arraystretch}{1.3}

 \begin{tabular}{|l|l|l|l|l||l|l|l|l|}
	 \hline
     & \multicolumn{4}{c|}{\textbf{$\theta\mbox{-}L_2$}} & \multicolumn{4}{c}{\textbf{$\beta\mbox{-}IE$}}\\
      \hline
	  \textbf{Model/Features} & \textbf{E} & \textbf{N} & \textbf{F} & \textbf{All} & \textbf{E} & \textbf{N} & \textbf{F} & \textbf{All} \\ \hline
Naive & 5.07 & 4.69 & 5.13 & 4.98 & 3.21 & 2.90 & 3.31 & 3.16 \\
GRU/T\textsubscript{CB} & 2.09 & 1.05 & 2.66 & \textbf{1.96} & 1.17 & 0.39 & 1.47 & \textbf{1.02} \\
Feed-forward/T\textsubscript{FT} & 2.95 & 1.96 & 3.13 & 2.70 & 1.71 & 1.03 & 1.84 & 1.55 \\
Feed-forward/T\textsubscript{Expert} & 3.58 & 3.29 & 4.21 & 3.75 & 2.16 & 1.95 & 2.63 & 2.28 \\
Random Forest/T\textsubscript{FT} & 3.04 & 2.21 & 3.70 & 3.06 & 1.80 & 1.25 & 2.23 & 1.81 \\
Random Forest/T\textsubscript{Expert} & 2.90 & 2.40 & 3.73 & 3.07 & 1.70 & 1.37 & 2.25 & 1.81 \\ \hline
Feed-forward/(GRU\_pred.+T\textsubscript{Expert})* & 2.52 & 1.76 & 3.06 & 2.48 & 1.42 & 0.90 & 1.8 & 1.40 \\
Random Forest/(GRU\_pred.+T\textsubscript{Expert})* & 1.71 & 0.60 & 2.50 & \textbf{1.66} & 1.04 & 0.28 & 1.43 & \textbf{0.94} \\ \hline
    \end{tabular}
    	\caption{OneShot text-level 
$\theta\mbox{-}L_2$ and $\beta\mbox{-}IE$ scores on test dataset (* indicates dependent model)}
\label{table:OneShot text-level proposed metrics} 
\end{table*}

\subsubsection{Sentence-level recommendations}
For the sentence level experiments, we employ a pretrained CamemBERT as a regression model by fine-tuning it with our training data. CamemBERT does not allow multi-value regression. Therefore,  this paper trains two CamemBERT models separately to recommend the lower bound (minimum age) and the upper bound (maximum age) of an age range for sentence level age recommendation. Later the mean age is derived from the recommended bounds. However, Feed-forward and Random Forest models are trained to directly recommend the age range like the OneShot age recommendation.

\begin{table*}[t!]
\centering
\small
	\renewcommand{\arraystretch}{1.3}

\begin{tabular}{|l|l|l|l|l||l|l|l|l|}
	 \hline
     & \multicolumn{4}{c|}{\textbf{Validation}} & \multicolumn{4}{c}{\textbf{Test}}\\
      \hline
	  \textbf{Model/Features} & \textbf{E} & \textbf{N} & \textbf{F} & \textbf{All} & \textbf{E} & \textbf{N} & \textbf{F} & \textbf{All} \\ \hline
Naive & 2.67 & 2.61 & 2.94 & 2.82 & 3.26 & 2.63 & 2.20 & 2.50 \\
CamemBERT/- & 1.93 & 1.3 & 2.21 & \textbf{2.00} & 2.02 & 1.28 & 1.94 & \textbf{1.83}\\
Feed-forward/S\textsubscript{FT} & 2.31 & 1.65 & 2.86 & 2.53 & 2.41 & 1.61 & 2.38 & 2.24 \\
Feed-forward/S\textsubscript{Expert} & 2.75 & 2.60 & 2.94 & 2.84 & 2.75 & 2.51 & 2.68 & 2.66 \\
Feed-forward/(S\textsubscript{FT}$\oplus$S\textsubscript{Expert}) & 2.19 & 1.60 & 2.87 & 2.49 & 2.31 & 1.55 & 2.39 & 2.21 \\
Random Forest/S\textsubscript{FT} & 2.42 & 2.15 & 2.43 & 2.39  & 2.68 & 2.02 & 2.01 & 2.15 \\
Random Forest/S\textsubscript{Expert} & 2.41 & 2.23 & 2.41 & 2.39 & 2.62 & 2.17 & 2.04 & 2.19 \\ \hline
Feed-forward/(CamemBERT\_pred.$\oplus$S\textsubscript{Expert})* & 2.08 & 1.38 & 2.71 & 2.34 & 2.08 & 1.34 & 2.33 & 2.09 \\ 
Random Forest/(CamemBERT\_pred.$\oplus$S\textsubscript{Expert})* & 1.97 & 1.22 & 2.42 & \textbf{2.12} & 2.00 & 1.18 & 1.98 & \textbf{1.83} \\
\hline
    \end{tabular}
    	\caption{Sentence level $\mu{}E$ scores (* indicates dependent model)}
\label{table:Sentence level muE} 
\end{table*}

\begin{table*}[t!]
\centering
\small    

	\renewcommand{\arraystretch}{1.3}

\begin{tabular}{|l|l|l|l|l||l|l|l|l|}
	 \hline
     & \multicolumn{4}{c|}{\textbf{$\theta\mbox{-}L_2$}} & \multicolumn{4}{c}{\textbf{$\beta\mbox{-}IE$}}\\
      \hline
	  \textbf{Model/Features} & \textbf{E} & \textbf{N} & \textbf{F} & \textbf{All} & \textbf{E} & \textbf{N} & \textbf{F} & \textbf{All} \\ \hline
Naive & 5.11 & 4.28 & 3.83 & 4.18 & 3.22 & 2.59 & 2.35 & 2.58 \\
CamemBERT/- & 3.49 & 2.44 & 3.41 & \textbf{3.25} & 2.13 & 1.38 & 2.03 & \textbf{1.93} \\
Feed-forward/S\textsubscript{FT} & 3.95 & 2.81 & 4.00 & 3.77 & 2.40 & 1.66 & 2.47 & 2.30\\
Feed-forward/S\textsubscript{Expert} & 4.42 & 4.11 & 4.46 & 4.39 & 2.73 & 2.54 & 2.79 & 2.73\\
Feed-forward/(S\textsubscript{FT}$\oplus$S\textsubscript{Expert}) & 3.80 & 2.72 & 4.01 & 3.72 & 2.31 & 1.60 & 2.48 & 2.28\\
Random Forest/S\textsubscript{FT} & 4.29 & 3.47 & 3.48 & 3.65 & 2.66 & 2.08 & 2.09 & 2.20 \\
Random Forest/S\textsubscript{Expert} & 4.22 & 3.67 & 3.56 & 3.72 & 2.62 & 2.21 & 2.14 & 2.25 \\ \hline
Feed-forward/(CamemBERT\_pred.$\oplus$S\textsubscript{Expert})* & 3.42 & 2.42 & 3.93 & 3.54 & 2.08 & 1.39 & 2.42 & 2.16 \\
Random Forest/(CamemBERT\_pred.$\oplus$S\textsubscript{Expert})* & 3.28 & 2.16 & 3.41 & \textbf{3.15} & 1.99 & 1.25 & 2.05 & \textbf{1.89}\\
 \hline
    \end{tabular}
\caption{Sentence-level 
$\theta\mbox{-}L_2$ and $\beta\mbox{-}IE$ scores on test dataset (* indicates dependent model)}
\label{table:Sentence-level proposed metrics} 
\end{table*}

Table \ref{table:Sentence level muE} details the sentence level $\mu{}E$ scores for the three genres on the validation and test datasets.
In this case, Feed-forward with $S_{Expert}$ features performs worst having a $\mu{}E$ score higher than the Naive approach. All other models perform better than the Naive one. 
The CamemBERT model with its contextual word embeddings achieves the best scores for each genre on both datasets. 
Interestingly, on Feed-forward model, $S_{FT}$ features perform better than the $S_{Expert}$ features. The $S_{FT}$ and $S_{Expert}$ features jointly improves a little over $S_{Expert}$ features but seems not significant. Like the OneShot text-level, Random Forest with $S_{Expert}$ features gets better score than the Feed-forward one. 
However, as dependent models (CamemBERT recommended min-age and max-age, concatenated with $S_{Expert}$ features), the Feed-forward and Random Forest do not improve the scores over the CamemBERT. 

We also observe the $\theta\mbox{-}L_2$ $\beta\mbox{-}IE$ scores (Table \ref{table:Sentence-level proposed metrics}) for the sentence level age recommendation on the test dataset and notice the coherency of the performances by different models as seen for $\mu{}E$ scores.
These scores show that CamemBERT model significantly outperforms the other models without requiring any explicit feature engineering. 
However, it is not surprising that the sentence level scores are worse than the OneShot text-level scores as the training sentences are not annotated by the experts rather it is done based on the text annotations which leads to inaccurate sentence annotation. 

\subsubsection{From sentences to text-level recommendations}

The sentence level scores still look very decent and it motivates us to explore aggregation of sentence level recommendations to make a text-level global recommendation. Here we study the mean aggregation which derives the mean value from the recommended ages of the sentences in a text.

\begin{table*}[t!]
\centering
\small
    \renewcommand{\arraystretch}{1.3}

    \begin{tabular}{|l|l|l|l|l||l|l|l|l|}
     \hline
     & \multicolumn{4}{c|}{\textbf{Validation}} & \multicolumn{4}{c}{\textbf{Test}}\\
      \hline
      \textbf{Model/Features} & \textbf{E} & \textbf{N} & \textbf{F} & \textbf{All} & \textbf{E} & \textbf{N} & \textbf{F} & \textbf{All} \\ \hline
Naive & 2.80 & 2.94 & 3.78 & 3.25 & 3.24 & 2.94 & 3.12 & 3.10 \\
CamemBERT/- & 1.86 & 1.12 & 1.94 & \textbf{1.72} & 1.92 & 1.16 & 1.73 & \textbf{1.59} \\
Feed-forward/S\textsubscript{FT} & 2.11 & 1.33 & 2.48 & 2.09 & 2.21 & 1.36 & 1.91 & 1.80 \\ 
Feed-forward/S\textsubscript{Expert} & 2.31 & 2.12 & 2.83 & 2.50 & 2.46 & 2.08 & 2.41 & 2.32 \\ 
Feed-forward/(S\textsubscript{FT}$\oplus$S\textsubscript{Expert}) & 2.01 & 1.29 & 2.47 & 2.04 & 2.12 & 1.30 & 1.93 & 1.76 \\
Random Forest/S\textsubscript{FT} & 2.45 & 2.14 & 2.84 & 2.55 & 2.67 & 2.07 & 2.42 & 2.37 \\
Random Forest/S\textsubscript{Expert} & 2.44 & 2.23 & 2.93 & 2.60 & 2.60 & 2.21 & 2.53 & 2.44 \\ \hline
    \end{tabular}
\caption{Text level $\mu{}E$ scores by Mean-aggregation}
\label{table:Text level muE by Mean-Aggregation} 
\end{table*}
%
\begin{table*}[t!]
\centering
\small
	\renewcommand{\arraystretch}{1.3}
 \begin{tabular}{|l|l|l|l|l||l|l|l|l|}
	 \hline
     & \multicolumn{4}{c|}{\textbf{$\theta\mbox{-}L_2$}} & \multicolumn{4}{c}{\textbf{$\beta\mbox{-}IE$}}\\
      \hline
	  \textbf{Model/Features} & \textbf{E} & \textbf{N} & \textbf{F} & \textbf{All} & \textbf{E} & \textbf{N} & \textbf{F} & \textbf{All} \\ \hline
Naive & 5.07 & 4.69 & 5.13 & 4.98 & 3.21 & 2.90 & 3.31 & 3.16 \\
CamemBERT/- & 3.27 & 2.18 & 3.07 & \textbf{2.83} & 1.97 & 1.21 & 1.78 & \textbf{1.64}\\
Feed-forward/S\textsubscript{FT} & 3.61 & 2.42 & 3.35 & 3.10 & 2.19 & 1.37 & 1.96 & 1.82\\
Feed-forward/Expert & 3.97 & 3.48 & 4.08 & 3.86 & 2.42 & 2.06 & 2.49 & 2.34\\
Feed-forward/(S\textsubscript{FT}$\oplus$S\textsubscript{Expert}) & 3.49 & 2.33 & 3.37 & 3.05 & 2.10 & 1.31 & 1.98 & 1.78\\
Random Forest/S\textsubscript{FT} & 4.26 & 3.46 & 4.08 & 3.92 & 2.63 & 2.04 & 2.48 & 2.38 \\
Random Forest/S\textsubscript{Expert} & 4.17 & 3.65 & 4.28 & 4.05 & 2.56 & 2.18 & 2.63 & 2.46\\ 
 \hline
    \end{tabular}
\caption{Text level 
$\theta\mbox{-}L_2$ and $\beta\mbox{-}IE$ scores by mean-aggregation on test dataset 
}
\label{table:mean-aggregation proposed metrics} 
\end{table*}

As shown in Table ~\ref{table:Text level muE by Mean-Aggregation}, by significantly outperforming the Naive approach, the mean-aggregation (especially for the CamemBERT model) gives very good $\mu{}E$ scores to recommend age for a text. CamemBERT obtains $\mu{}E$ scores of $1.72$ and $1.59$ on the validation and test datasets which are $0.33$ and $0.61$ points worse than the best (GRU/T\textsubscript{CB}) OneShot model. Although, Feed-forward performs better than Random Forest, both get some good scores on the validation and test datasets. Like the OneShot text-level and sentence-level, here again we notice that on the Feed-forward model, $S_{FT}$ features performs better than the $S_{Expert}$ features. Jointly the $S_{FT}$ and $S_{Expert}$ features gets insignificantly a little better score over $S_{FT}$ features alone. 
Similar behaviour is noticed with the $\theta\mbox{-}L_2$ and $\beta\mbox{-}IE$ scores on the test dataset as shown in Table~\ref{table:mean-aggregation proposed metrics}.  

Overall, the scores by mean-aggregation are very promising and this aggregation method provides significant insights on the age-recommendation particularly for the authors view-point. This approach can be very effective for an author to incrementally verify a text, either it is suitable for a target age.

\subsubsection{Comparison of recommendations per age and age-range}
Up to now, [GRU/T\textsubscript{CB}, Feed-forward/T\textsubscript{FT}] and [CamemBERT/-, Feed-forward/S\textsubscript{FT}] are the best performing two pairs of models for the OneShot text-level and sentence-level age recommendation. Here we compare their recommendation performances for each age and age-range.

\begin{figure}[t]
    \begin{center}
    \includegraphics[width=0.8\linewidth]{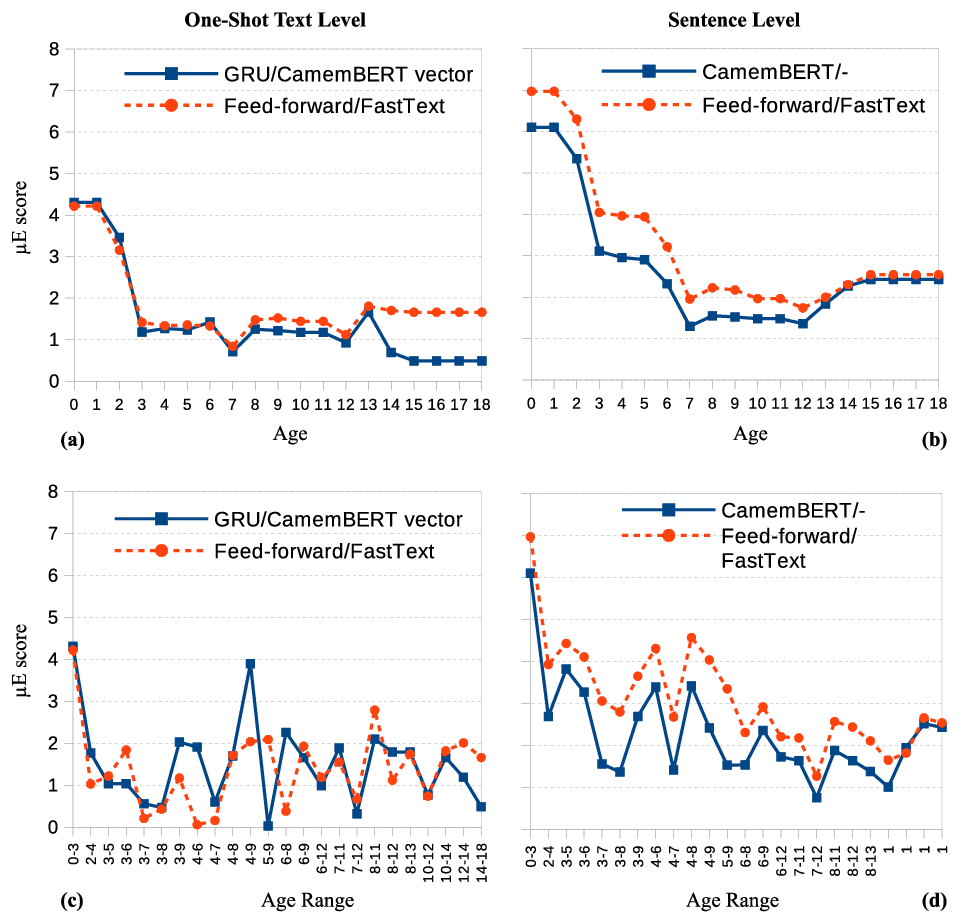}
    \caption{\label{fig:score_charts} Per age (a and b), and per age-range (c and d) $\mu{}E$ scores at the OneShot text level (a and c) and sentence level (b and d) on test set}
    \end{center}
\end{figure}

Fig.~\ref{fig:score_charts} shows the $\mu{}E$ scores per age (a and b), and per age-range (c and d) at the text level (a and c) and sentence level (b and d) on test set. Almost in all cases (except text level scores per age-range), the GRU/T\textsubscript{CB} (text level) and CamemBERT/- (sentence level) outperform the Feed-forward/S\textsubscript{FT} (text level) and Feed-forward/S\textsubscript{FT} (sentence level) models.

For OneShot text-level age recommendation, GRU/T\textsubscript{CB} performs better for recommending ages of 8 to 12 years (older children) and 14 to 18 years (adults). For other ages, the two models perform quite similar. However, in sentence level age recommendation, CamemBERT/- gets much better scores for child-ages than adult-ages. 
Overall, CamemBERT has outstanding contributions in age recommendation task.


\subsection{Comparison with Experts}
\label{sec:experts}

This section compares the recommendations from our best models with those of three psycho-linguist experts on sentences and texts. To do so, the experts were requested to annotate $20$ texts from the test set and $80$ sentences randomly selected from these $20$ texts. The different models are applied on the same sentences and texts. Table~\ref{table:expert_test_scores} compares $\mu{}E$, $\theta\mbox{-}L_2$ and $\beta\mbox{-}IE$ scores measured for the Naive approach, CamemBERT and GRU models, every expert, and when averaging the expert's recommendations. 

\begin{table*}[t!]
\centering
\small
	\renewcommand{\arraystretch}{1.3}
 \begin{tabular}{|l|l|l|l||l|l|l|l|}
	 \hline
\multicolumn{4}{|c||}{\textbf{Sentence Level Prediction}} & \multicolumn{4}{c|}{\textbf{Text Level Prediction}}\\ \hline
\textbf{Model} & \textbf{$\mu{}E$} & \textbf{$\theta\mbox{-}L_2$} & \textbf{$\beta\mbox{-}IE$} & \textbf{Model}  & \textbf{$\mu{}E$} & \textbf{$\theta\mbox{-}L_2$} & \textbf{$\beta\mbox{-}IE$} \\
\hline
Naive & 5.02 & 7.70 & 5.14 & Naive & 4.55 & 7.11 & 4.74 \\
\multirow{2}{*}{CamemBERT/-} & \multirow{2}{*}{3.43} &  \multirow{2}{*}{5.54} & \multirow{2}{*}{3.21} & Mean Aggr. (CamemBERT/-) & 2.95 & 4.83 & 2.76 \\
 & & & & OneShot (GRU/T\textsubscript{CB}) & \textbf{1.47} & \textbf{2.72} & \textbf{1.43} \\
\hline 
Expert 1 & 3.28 & 5.23 & 3.27 & \multicolumn{1}{l|}{Expert 1} & 2.60 & 4.23 & 2.60 \\ 
Expert 2 & 3.25 & 5.28 & 3.33 & \multicolumn{1}{l|}{Expert 2} & 3.50 & 5.76 & 3.63 \\
Expert 3 & 3.18 & 5.28 & 3.26 & \multicolumn{1}{l|}{Expert 3} & 2.72 & 4.47 & 2.70 \\
\hline
Expert mean & \textbf{2.90} & \textbf{4.76} & \textbf{2.93} & \multicolumn{1}{l|}{Expert mean} & 2.95 & 4.81 & 2.95 \\
\hline
    \end{tabular}
\caption{Age prediction scores ($\mu{}E$, $\theta\mbox{-}L_2$ and $\beta\mbox{-}IE$) for the expert annotated 80 sentences and 24 texts from the test set.}
\label{table:expert_test_scores} 
\end{table*}
%
\begin{table*}[t!]
\centering
\small
	\renewcommand{\arraystretch}{1.3}
\begin{tabular}{|l|l|l|l|l|l|}
	 \hline
	  \textbf{Sentence} & \makecell[c]{\textbf{Actual Age} \\ \textbf{Range / Mean}} & \makecell[c]{\textbf{CamemBERT}\\ \textbf{Range / Mean}} & \textbf{$\mu{}E$} & \textbf{$\theta\mbox{-}L_2$} & \textbf{$\beta\mbox{-}IE$} \\ 
	  \hline

\makecell[l]{\textbf{1.} 
Yum, here is a Mosquito!} & [4, 8] / 6.0 & \textbf{[4.73, 7.39] / 6.06} & 0.06 & 0.95 & 0.55 \\ 
\hdashline
\makecell[l]{\textbf{2.} This afternoon, Tremolo was to\\ play at a funeral.} & [8, 11] / 9.5  & \textbf{[8.98, 11.27] / 10.13} & 0.63 & 1.27 & 0.60 \\
\hdashline
\makecell[l]{\textbf{3.} If we wait later, it is more complicated.} & [12, 14] / 13.0 & [7.80, 12.04] / 9.92 & 3.08 & 5.30 & 3.62 \\
\hdashline
\makecell[l]{\textbf{4.} By then, you will have had time\\ to comfort a little girl who has already\\ shed tears on her fault.} & [14, 18] / 16.0 &  [7.35, 11.50] / 9.43 & 6.57 & 9.80 & 6.60 \\
      \hline
    \end{tabular}
\caption{Examples of age recommendation by the CamemBERT/- model for some expert-annotated sentences. Here the sentences are translated in English while the French version is available in Table~\ref{table:example_preds_fr}, Appendix~\ref{app:results}.}
\label{table:example_preds} 
\end{table*}

The CamemBERT and GRU models significantly outperform the Naive approach for the sentence-level and text-level age recommendation. However, CamemBERT/- model gets high $\mu{}E$, $\theta\mbox{-}L_2$ and $\beta\mbox{-}IE$ scores at the sentence level but still performs quite similar to the individual experts. 
At the text level, CamemBERT/- with mean-aggregation obtains the same $\mu{}E$ as that of the expert-mean. As an OneShot model, using sentence-vectors from CamemBERT, GRU/T\textsubscript{CB} significantly outperforms the experts. These results indicate that the models built on CamemBERT perform better than human annotators. 
Table~\ref{table:example_preds} presents some sentence level age recommendations where the sentences are annotated by experts. The CamemBERT/- model gives very good recommendation for the first two sentences while it fails very badly for the last two instances.





%% file: section_7_feature_analysis_EN_v3.tex
\section{Feature Analysis for Explainability}
\label{sec:feature_analysis}

Overall the age recommendation models obtain very decent scores. Although the \textit{Expert} feature set (with any regressor) does not achieve the best score, it allows us to analyze which linguistic properties have higher contributions in age recommendation. As an opening of explainability of the age prediction models, this section explores feature ranking by different approaches. Before detail analysis, Table~\ref{table:def_expert_features} describes some linguistic expert features which are presented in this section\footnote{
Appendix~\ref{app:expert_features} describes all the expert features.}.

\begin{table*}
\centering
\small
	\renewcommand{\arraystretch}{1.3}
\begin{tabular}{|r|l|}
    \hline
    \textbf{Feature Name} & \textbf{Description} \\ \hline
    SentenceLength & Number of words in a sentence. \\
    WordLengthStd & Std. deviation of the length of each word in a sentence. \\
    LemmaDiversity & Number of different lemmas in a sentence. \\
    StopwordsProportion & Proportion of stop words to the total number of words in a sentence.\\
    PhonemeNumberSentence & Number of phonemes in the sentence (phonetic length). \\
    PhonemeDiversitySentence & Number of different phonemes in a sentence. \\
    PhonemeNumberAvg & Average number of phonemes per word. \\
    PhonemeNumberStd & Std. deviation of the number of phonemes per word. \\
    PhonemeProbStd & Std. deviation of the average probability of the phonemes in a sentence. \\
    PhonemeDiversityStd & Std. deviation of the number of different phonemes in each word. \\
    PolarityScore & Polarity score of a sentence [-1, 1]. \\
    \hline
\end{tabular}
    \caption{Description of some the linguistic expert features}
    \label{table:def_expert_features}
\end{table*}

\subsection{Correlation Between Feature and Age}
At this point, we are interested to measure the correlation (Pearson correlation coefficient) between the features and ages (on test set) to identify the most contributory features. 

\begin{figure}[t!]
\begin{center}
\includegraphics[width=0.60\linewidth]{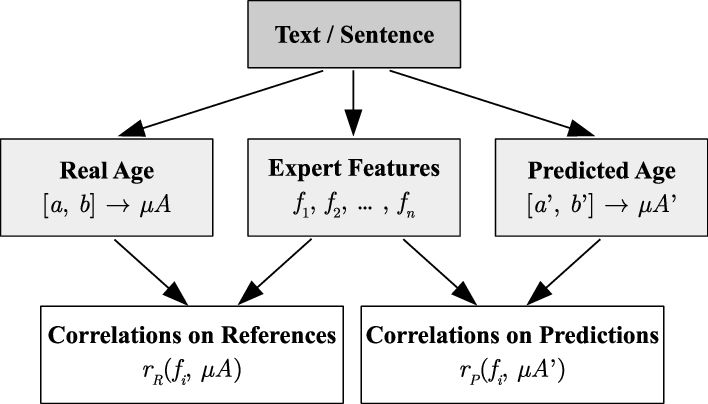}
\caption{Correlation between expert feature and real age, and expert feature and predicted age}
\label{fig:correlation}
\end{center}
\end{figure}
%
\begin{table*}[t!]
\centering
\small
	\renewcommand{\arraystretch}{1.3}
 \begin{tabular}{{|l|l|l|l||l|l|l|}}
	 \hline
     & \multicolumn{3}{c|}{\textbf{Real Age}} & \multicolumn{3}{c}{\textbf{Predicted Age}}\\
      \hline
	  \textbf{Features} & \textbf{Text} & \textbf{Sentence} & \textbf{Avg.*} & \makecell[c]{\textbf{Text} \\ \textbf{(GRU/T\textsubscript{CB})}} & \makecell[c]{\textbf{Sentence} \\ \textbf{(CamemBERT/-)}} & \textbf{Avg.} \\ \hline
PhonemeNumberAvg & 0.56 / 2 & 0.25 / 1 & 0.40 / 1 & 0.48 / 3 & 0.27 / 1 & 0.38 / 2 \\
PhonemeNumberStd & 0.58 / 1 & 0.22 / 2 & 0.40 / 2 & 0.52 / 1 & 0.26 / 2 & 0.39 / 1 \\
WordLengthStd & 0.54 / 3 & 0.19 / 7 & 0.37 / 3 & 0.50 / 2 & 0.25 / 4 & 0.38 / 3 \\
LemmaDiversity & 0.50 / 4 & 0.22 / 4 & 0.36 / 4 & 0.48 / 6 & 0.24 / 5 & 0.36 / 5 \\
SentenceLength & 0.49 / 5 & 0.22 / 3 & 0.36 / 5 & 0.48 / 4 & 0.25 / 3 & 0.37 / 4 \\
\hline
\hline
PhonemeNumberSentence  &  -0.34 / 2  &  -0.15 / 1  &  -0.25 / 1  &  -0.31 / 3  &  -0.16 / 1  & -0.24 / 2 \\
PhonemeDiversitySentence  &  -0.34 / 3  &  -0.15 / 2  &  -0.25 / 2  &  -0.31 / 4  &  -0.16 / 2  & -0.24 / 3 \\
PhonemeProbStd  &  -0.37 / 1  &  -0.10 / 4  &  -0.24 / 3  &  -0.39 / 1  &  -0.15 / 3  & -0.27 / 1 \\
PhonemesDiversityStd  &  -0.32 / 6  &  -0.14 / 3  &  -0.23 / 4  &  -0.25 / 6  &  -0.15 / 4  & -0.20 / 4 \\
StopWordsProportion  &  -0.33 / 5  &  -0.08 / 5  &  -0.20 / 5  &  -0.31 / 5  &  -0.09 / 6  & -0.20 / 5 \\
    \hline
    \end{tabular}
	\caption{Correlation (correlation score / feature rank) of \textit{Expert} features to the real and predicted mean-ages at the text and sentence levels}
\label{table:correlation} 
\end{table*}

First, correlation is measured between each expert-feature and the real ages at the sentence and text levels, and then done similar with the predicted ages from two models, GRU/T\textsubscript{CB} (text level) and CamemBERT/- (sentence level) as shown in Fig.~\ref{fig:correlation}. 
We hypothesize that the top ranked correlated features will be similar for the real age and recommended age for a good age recommendation model. 
Table~\ref{table:correlation} shows the top 5 positively correlated (upper part) and top 5 negatively correlated (bottom part) features for text level and sentence level predictions. Each correlation score is presented with its rank among the features. For example, in second column (Real Age/Text), the correlation score for \textit{PhonemeNumberStd} (0.58 / 1) is 0.58 with the rank of 1. From this table, it is very clear that the top ranked positively and negatively correlated features are common in all settings although their ranks displace a little. 
From the correlation based feature ranking, one can notice that among different types of features the \textit{phonetic} features obtains the best rank scores.
\subsection{Permutation Feature Ranking}
The idea of permutation feature ranking is to randomly shuffle an individual feature value to break the relationship between the feature and the target value. Thus it measures the drop of performance due to discarding the feature, and finally ranks the features. Here we do it for the expert features (T\textsubscript{Expert} and S\textsubscript{Expert}) at the sentence level and text level on the test set by employing Random Forest regressor, a built-in library in scikit-learn. 

\begin{table}[t!]
\centering
\small
	\renewcommand{\arraystretch}{1.3}
 \begin{tabular}{|l|l|l|l|}
      \hline
      \multicolumn{2}{|c|}{\textbf{Sentence Level}} & \multicolumn{2}{c|}{\textbf{Text Level}}\\
      \hline
    \textbf{Feature} & \textbf{Score} & \textbf{Feature} & \textbf{Score} \\ \hline

PhonemeNumberAvg & 0.1021 & PhonemeNumberAvg & 0.1596 \\
PhonemeNumberStd & 0.0383 & SentenceLength & 0.1425 \\
SentenceLength & 0.0368 & PhonemeNumberStd & 0.0847 \\
PhonemeProbStd & 0.0132 & PhonemeProbStd & 0.0245 \\
StopWordsProportion & 0.0130 & PolarityScore & 0.0213 \\
     \hline
    \end{tabular}
	\caption{Top 5 \textit{Expert} features (with rank-scores) by permutation feature ranking at the sentence and text levels}
\label{table:permutation feature ranks} 
\end{table}

Table~\ref{table:permutation feature ranks} enlists the top 5 features for different feature combinations at the sentence and text levels. For ranking on the \textit{Expert} features, most of the top ranked features are related to \textit{Phonetics}. Moreover, the \textit{SentenceLength}
and \textit{StopWordsProportion} are found very important features for age recommendation at the sentence and text levels. These features are also seen in the list of top ranked features by correlation based feature ranking in Table~\ref{table:correlation}.

\begin{table}[t!]
\centering
\small
 \begin{tabular}{|l|l|l|}
	 \hline
	  \textbf{Features} & \textbf{Sentence} & \textbf{Text} \\ \hline
\textbf{Expert} & \textbf{2.66} & \textbf{2.22} \\ \hdashline
Expert - Phonetics & 2.77 / +0.11 & 2.33 / +0.11 \\
Expert - Dependencies & 2.71 / +0.05 & 2.27 / +0.05 \\
Expert - Morphosyntax & 2.71 / +0.05 & 2.26 / +0.04 \\
Expert - Lexicon & 2.69 / +0.03 & 2.29 / +0.07 \\
\hdashline
Expert - Sentiments & 2.69 / +0.03 & 2.18 / -0.04 \\
Expert - Person/Number & 2.68 / +0.02 & 2.16 / -0.06 \\
Expert - Connectors & 2.67 / +0.01 & 2.21 / -0.01 \\
Expert - VerbalTenses & 2.66 / +0.00 & 2.19 / -0.03 \\
Expert - Graphemes & 2.65 / -0.01 & 2.18 / -0.04 \\
\hline
    \end{tabular}
\caption{\label{table:Ablation test muE} Ablation test ($\mu{}E$ / $\Delta$) of the \textit{Expert} features with Feed-forward model at the sentence and text levels}
\end{table}

\subsection{Ablation Test}
Finally we do an ablation test with different categories of the \textit{Expert} features on the test set. The idea of ablation test is like the permutation feature ranking where an individual feature is shuffled and its impact on the performance is measured. Here ablation test is done with the Feed-forward model,
but instead of shuffling a single feature, all the features of same category are discarded.

Table~\ref{table:Ablation test muE} presents the $\mu{}E$ scores of the ablation test where almost all the feature categories except \textit{VerbalTenses} and \textit{Graphemes} show positive impacts at sentence level. Interestingly, the \textit{Phonetics, Dependencies, Morphosyntax} and \textit{Lexicon} features have positive impacts at both sentence and text levels. All the feature ranking approaches studied in this paper find the phonetics 
features the most common and contributory in any settings. Although these feature categories individually do not have very high impact, they altogether perform quite well in age recommendation task.

%% file: section_8_conclusion.tex
\section{Conclusion and Future Work}
\label{sec:conclusion}
To recommend adequate texts to the children of different ages, this paper presented a detail work on age prediction from texts/sentences. This study discussed several aspects about the definition of target age, evaluation metrics, and explored various machine learning models for age recommendation. Almost all the models perform very decently while particularly two models built on top of CamemBERT achieve the best scores, even by outperforming the human experts. The results at the text level are very good and suggest the use of the best model for real applications. On sentences, the error rates are higher, probably due to less reliable references (derived from the texts). 
Our preliminary study on the explainability of age recommendation models finds that the linguistic phonetic features are most prominent regarding the different difficulty levels of the texts/sentences. The dependency, morphosyntax and lexicon features also show their importance in some extents. 

As an explainability task, to explain text-difficulty for different child-ages, this paper looks forward 
to identifying the potential words and phrases locally (instead of global context) which impact the difficulty levels in a text/sentence. In future, we also aim to work on text simplification (by paraphrasing) to offer a better reading experience to the children.

%% file: appendix_metrics.tex
\section{Details of the Evaluation Metrics}
\label{app:metrics}

\subsection{Combining distance and angle}

\begin{figure}[ht]
 \begin{center}
 \includegraphics[scale=0.9]{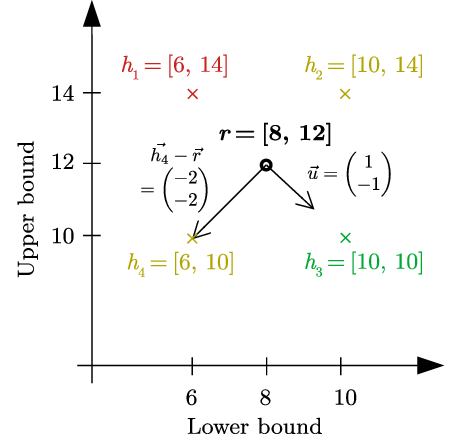}
   \caption{\label{fig:range_metrics_geometry}Examples of 4 ranges with equal $L_2$ errors.}
 \end{center}
\end{figure}

As depicted in Figure~\ref{fig:range_metrics_geometry}, different hypotheses can lead to a same error, whereas one may expect to distinguish them. Especially when introducing the uncertainty of the reference interval, one may consider that inner errors are less severe than outer ones. For instance, based on the reference $[8, 12]$, $h_3 = [10, 10]$ is a better recommendation than $h_1 = [6, 14]$. Considering the cosine of the angle $\theta$ between of the vector $\vec{h}-\vec{r} = \twodvector{c - a}{d - b}$ and the inner direction $\vec{\imath} = \twodvector{1}{-1}$ alleviates this situation. The vector $\vec{\imath}$ symbolizes the fact that it is fine if the hypothesized lower bound is greater than the one of the reference, or if the hypothesized upper bound is lower than the one of the reference. An ideal value for $\theta$ is $0$ (cosine is $1$) while the worst situation is $\theta = \pi$ or $-\pi$ (cosine is $-1$).

\subsection{Error values for each metric}

Table~\ref{tab:range_metrics_values} reports the error given by various metrics over the set of 20~samples considered in Section~\ref{sec:metrics_experiments}, sorted according to their ascending Oracle rank.

\begin{table}[t!]
\centering
\small
\includegraphics[scale=0.85]{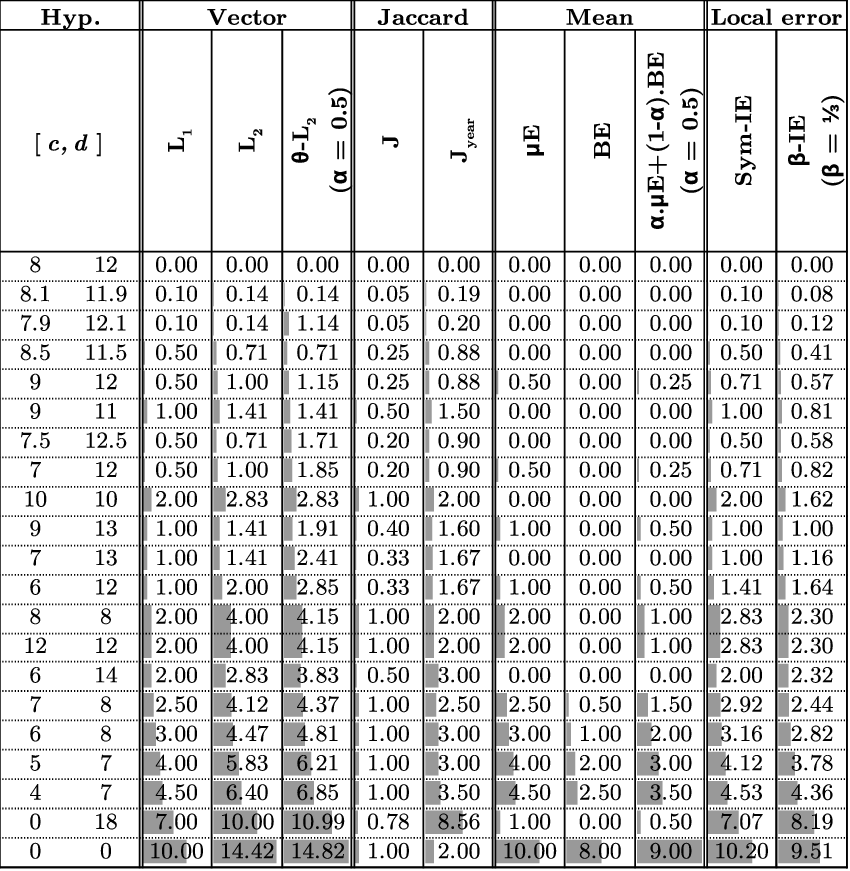}
\caption{\label{tab:range_metrics_values} Values returned by the metrics for various hypothesis intervals $[c, d]$ according to the reference interval $[a, b] = [8, 12]$.}
\end{table}

%% file: appendix_expert_features.tex
\section{Expert Features}
\label{app:expert_features}

The state of the art leads us to consider a list of 38 linguistic aspects for age recommendation task. As detailed below, these aspects are gathered in 9 categories, totaling a feature vector of 107 real values for each input text/sentences.

\begin{enumerate}
\item \textbf{Lexical information (5 features)}
    \begin{itemize}
    \item Mean and standard deviation of log probability of the words in French. Log probabilities have been derived from the language model for French for the speech recognition, trained on types of various types.
    \item Diversity of lemmas.
    \item Mean and standard deviation over the frequencies of the words.
    \end{itemize}

\item \textbf{Graphy/typography (6 features)}
    \begin{itemize}
    \item Mean and standard deviation over the graphical confusability score of the words. To do so, we consider a graphical confusion score c(x, y) between two graphemes x and y. Then, given a word
    $w = [w_1, w_2 ... w_N ]$, the confusability score if computed as the cumulative confusion between each pair of consecutive graphemes in the word, that is: $\sum_{i=1}^{N-1} c(w_i, w_{i+1} )$. In practice, the confusion score c is taken from~\cite{Geyer1977}.
    \item Mean and standard deviation over of the length of the words.
    \item Ratio of characters (including punctuation marks) against the number of words.
    \item Ratio of punctuation marks against the number of words.
    \end{itemize}

\item \textbf{Morphosyntax (7 features)}
    \begin{itemize}
    \item Proportion of the following grammatical classes:
    verbs, state verbs, names, adjectives, clitics, temporal adverbs. Part of speech tags are generated using Bonsai~\cite{Candito}.
    \item Proportion of stop words in a list of 114 words from~\cite{ranks2019}.
    \end{itemize}

\item \textbf{Verbal tenses (24 features)}
    \begin{itemize}
    \item Number of different verbal tenses
    \item Proportions of 7 so-called simple tenses: present, past simple, future, imperfect, subjunctive present, conditional present, infinitive.
    \item Proportions of 7 composed tenses: compound past (passé composé), past past (passé antérieur), future past (futur antérieur), more than perfect (plus que parfait), subjunctive past, past conditional, past infinitive.
    \item Number of different temporal systems: past, present, future.
    \item Proportions of conjugated verbs for each of the 3 temporal systems: past, present, future.
    \item Proportion of compound tenses.
    \item Proportion of simple tenses.
    \item Proportion of each mode: infinitive, indicative, subjunctive.
    \end{itemize}

\item \textbf{Genders and numbers (5 features)}
    \begin{itemize}
    \item Proportion of conjugated verbs in the first person.
    \item Proportion of conjugated verbs in the second person.
    \item Proportion of conjugated verbs in the third person.
    \item Proportion of conjugated verbs in the singular form.
    \item Proportion of conjugated verbs in the plural form.
    \end{itemize}

\item \textbf{Syntactic dependencies (8 features)}
    \begin{itemize}
    \item Number of words per sentence.
    \item Average distances (word count) between a word and its dependencies. Dependency parsing is achieved using Bonsai~\cite{Candito}.
    \item Maximum distances (word count) between a word and its dependencies.
    \item Mean and standard deviation of dependencies per word (words that points to a given word).
    \item Mean and standard deviation of the distances between each word and the words to which it points.
    \item Depth of the dependency tree.
    \end{itemize}

\item \textbf{Logical connectors (16 features)}
    \begin{itemize}
    \item Proportion of logical connectors for each of the following types: addition; time; goal; cause; comparison; concession; conclusion; condition; consequence; enumeration; explanation; illustration; justification; opposition; restriction; exclusion. Since the way to gather connectors in categories, varies across papers, the categorization used is a consensus of all of them.
    \end{itemize}

\item \textbf{Phonetics (9 features)}
    \begin{itemize}
    \item Number of phonemes in the sentence, as generated using the grapheme-to-phoneme convertor of eSpeak~\cite{eSpeak}.
    \item Number of different phonemes in the sentence.
    \item Frequency of the phonemes over the whole sentence.
    \item Mean and variance of the phonetic ordinariness scores of the words. The ordinariness score is computed as the average probability of appearance of each phoneme in French, as given in~\cite{ConfPhonem}.
    \item Mean and variance of the word-based diversity of the phonemes.
    \item Mean and variance of the number of phonemes per word.
    \end{itemize}

\item \textbf{Sentiments/emotions (27 features)}
    \begin{itemize}
    \item Score of subjectivity as used in the sentiment classifier TextBlob~\cite{loria2018textblob}.
    \item Score of polarity (still using TextBlob).
    \item Proportion of words identified as trigger for a predefined set of 25 emotions: neutral, admiration, love, appeasement, daring, anger, behavior, guilt, disgust, displeasure, desire, embarrassment, empathy, pride, impassibility, inhumanity, jealousy, joy, contempt, unspecified, pride, fear, resentment, surprise, sadness.
    \end{itemize}
    This list is a refinement of the EMOTAIX dictionary~\cite{piolat2009example}.
\vspace{-2mm}
\end{enumerate}

%% file: appendix_results.tex

\section{Results}

\label{app:results}
\begin{table}[h!]
\centering
\small
	\renewcommand{\arraystretch}{1.5}
 \begin{tabular}{|l|l|l|l|l|l|}
	 \hline
	  \textbf{Sentence} & \makecell[c]{\textbf{Actual Age} \\ \textbf{Range / Mean}} & \makecell[c]{\textbf{CamemBERT}\\ \textbf{Range / Mean}} & \textbf{$\mu{}E$} & \textbf{$\theta\mbox{-}L_2$} & \textbf{$\beta\mbox{-}IE$} \\ 
	  \hline

\makecell[l]{\textbf{1.} 
Miam, voilà un moucheron!} & [4, 8] / 6.0 & \textbf{[4.73, 7.39] / 6.06} & 0.06 & 0.95 & 0.55 \\ 
\hdashline
\makecell[l]{\textbf{2.} Cet après-midi-là, Trémolo devait\\ jouer à un enterrement.} & [8, 11] / 9.5  & \textbf{[8.98, 11.27] / 10.13} & 0.62 & 1.27 & 0.60 \\
\hdashline
\makecell[l]{\textbf{3.} Si on attend plus tard, c'est plus\\ compliqué.} & [12, 14] / 13.0 & [7.80, 12.04] / 9.92 & 3.08 & 5.30 & 3.62 \\
\hdashline
\makecell[l]{\textbf{4.} D'ici là, vous aurez eu le temps de\\ rassurer une petite fille qui a déjà \\ versé des larmes sur sa faute.} & [14, 18] / 16.0 &  [7.35, 11.50] / 9.43 & 6.57 & 9.80 & 6.60 \\
      \hline
    \end{tabular}
\caption{Examples of age recommendation by the CamemBERT/- model for some expert-annotated sentences in French.}
\label{table:example_preds_fr} 
\end{table}


